\documentclass[conference]{IEEEtran}
\IEEEoverridecommandlockouts
\usepackage{cite}
\usepackage{float}
\usepackage{amsmath,amssymb,amsfonts}
\usepackage{algorithmic}
\usepackage{graphicx}
\usepackage{textcomp}
\usepackage{xcolor}
\usepackage{pgfplots}
\usepackage{colortbl}
\usepackage{booktabs}
\usepackage{adjustbox}
\pgfplotsset{compat=1.18}
\definecolor{low}{RGB}{255,255,255}   % white
\definecolor{mid}{RGB}{255,230,128}   % light yellow
\definecolor{high}{RGB}{255,100,80}
\usepackage{tabularx}
\usepackage{times}  % DO NOT CHANGE THIS
\usepackage{helvet}  % DO NOT CHANGE THIS
\usepackage{courier}  % DO NOT CHANGE THIS
\usepackage[hyphens]{url}  % DO NOT CHANGE THIS
\usepackage{graphicx} % DO NOT CHANGE THIS
\usepackage{xcolor}
\usepackage{tcolorbox}
\usepackage{caption} % DO NOT CHANGE THIS AND DO NOT ADD ANY OPTIONS TO IT
\usepackage{algorithm}
\usepackage{algorithmic}
\usepackage{amsmath}
\usepackage{amssymb}
\usepackage{array}
\usepackage{booktabs}
\usepackage{longtable}
\usepackage[utf8]{inputenc}
\usepackage{enumitem}
\usepackage{subcaption}  % defines the subtable environment
\usepackage{pdfpages}
\tcbuselibrary{breakable}
\usepackage{hyperref}

\usepackage{newfloat}
\usepackage{listings}
\DeclareCaptionStyle{ruled}{labelfont=normalfont,labelsep=colon,strut=off} % DO NOT CHANGE THIS
\floatstyle{ruled}
\newfloat{listing}{tb}{lst}{}
\floatname{listing}{Listing}
\newcommand{\prompt}{\raisebox{-0.2pt}{\includegraphics[scale=0.5]{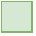}}}
\newcommand{\model}{\raisebox{-0.2pt}{\includegraphics[scale=0.5]{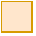}}}
\newcommand{\ann}{\raisebox{-0.2pt}{\includegraphics[scale=0.5]{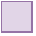}}}
\newcommand{\ds}{\raisebox{-0.2pt}{\includegraphics[scale=0.5]{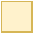}}}
\newcommand{\sen}{\raisebox{-0.2pt}{\includegraphics[scale=0.5]{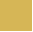}}}

\newcommand{\cause}{\raisebox{-0.2pt}{\includegraphics[scale=0.5]{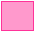}}}
\newcommand{\effect}{\raisebox{-0.2pt}{\includegraphics[scale=0.5]{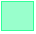}}}

\def\BibTeX{{\rm B\kern-.05em{\sc i\kern-.025em b}\kern-.08em
    T\kern-.1667em\lower.7ex\hbox{E}\kern-.125emX}}
\begin{document}

\title{Benchmarking LLMs for Pairwise Causal Discovery in Biomedical and Multi-Domain Contexts}

\author{
    \IEEEauthorblockN{
        Sydney Anuyah\textsuperscript{1},
        Sneha Shajee-Mohan\textsuperscript{1},
          Ankit-Singh Chauhan\textsuperscript{1},
        Sunandan Chakraborty\textsuperscript{1}
    }
    \IEEEauthorblockA{
        \textsuperscript{1}Indiana University, Indianapolis \\
        \ sanuyah@iu.edu, sshajee@iu.edu, ankichau@iu.edu, sunchak@iu.edu
    }
}

\maketitle

\begin{abstract}
The safe deployment of large language models (LLMs) in high-stakes fields like biomedicine, requires them to be able to reason about cause and effect. We investigate this ability by testing 13 open-source LLMs on a fundamental task: pairwise causal discovery (PCD) from text. Our benchmark, using 12 diverse datasets, evaluates two core skills: 1) \textbf{Causal Detection} (identifying if a text contains a causal link) and 2) \textbf{Causal Extraction} (pulling out the exact cause and effect phrases). We tested various prompting methods, from simple instructions (zero-shot) to more complex strategies like Chain-of-Thought (CoT) and Few-shot In-Context Learning (FICL).

The results show major deficiencies in current models. The best model for detection, DeepSeek-R1-Distill-Llama-70B, only achieved a mean score of 49.57\% ($C_{detect}$), while the best for extraction, Qwen2.5-Coder-32B-Instruct, reached just 47.12\% ($C_{extract}$). Models performed best on simple, explicit, single-sentence relations. However, performance plummeted for more difficult (and realistic) cases, such as implicit relationships, links spanning multiple sentences, and texts containing multiple causal pairs. We provide a unified evaluation framework, built on a dataset validated with high inter-annotator agreement ($\kappa \ge 0.758$), and make all our data, code, and prompts publicly available to spur further research. \href{https://github.com/sydneyanuyah/CausalDiscovery}{Code available here: https://github.com/sydneyanuyah/CausalDiscovery}
\end{abstract}

\section{Introduction}
Large language models (LLMs) are rapidly transforming healthcare. Recent applications show promise in data mining \cite{gopalakrishnan2023text}, personalized healthcare \cite{wan2025largelanguagemodelscausal, chi2024unveiling}, and drug discovery \cite{gopalakrishnan2024causality, jin2020inter}. This rapid transformation is enabling sophisticated analysis of unstructured data like clinical notes, medical literature, and electronic health records (EHRs) \cite{ma2025causalinferencelargelanguage, takayanagi2024chatgpt}. 
However, their integration into clinical settings faces critical challenges with reliability. For an LLM to be a trustworthy partner, it must move beyond pattern matching to exhibit genuine reasoning. A key aspect of this is distinguishing causation from correlation. An error in discerning whether a treatment \textit{caused} an improvement or was merely \textit{administered concurrently} could have severe consequences, rendering the technology unsafe for clinical decision support.
This challenge is formally addressed by the field of \textit{causal discovery (CD)}.

This fundamental challenge of inferring cause-and-effect is formally addressed by the field of \textit{causal discovery (CD)}. CD aims to infer causal relationships from data. It includes discovering the causal relationship graph and the structural equations associated with these causal relationships. Pairwise causal discovery (PCD) is a foundational problem in CD and it involves determining the cause and effect pairs typically from text data. This is crucial for medical applications where inferring a complete causal graph is impractical, but identifying individual links (e.g., ``medication A leads to side effect B'') provides immediate clinical value. This task is non-trivial; linguistic expressions of causality are nuanced, and simple lexical cues like ``causes" are often misleading, as shown in Table 1. To ensure validity, our work uses a strict annotation protocol, defining causality as a factual, unambiguous relationship stated in the text.

While LLMs offer a promising alternative to traditional CD methods, which struggle with unstructured text \cite{wan2024large}, their ability to perform PCD in an unsupervised, prompt-based manner remains underexplored \cite{wu2024causalitylargelanguagemodels}.
\begin{table}[t]
\small
Table 1: {Causal and Non-Causal Sentences in Practice}
\vspace{3pt}
% Use tabularx to make the table fit the column width
% The 'X' column type automatically wraps text

\begin{tabularx}{\columnwidth}{@{}lX@{}} 
    \toprule
    \textbf{Type} & \textbf{Example} \\
    \midrule
    No-Marker inter & The patient has an infection. His temperature rose. \\
    Marker inter & A sedative was given. As a result, the patient slept. \\
    Marker intra & The fracture was caused by the fall. \\
    No-Marker intra & Smoking damages the lungs. \\
    Marker, not truly causal & The article links coffee to longevity. \\
    Marker, not causal & The cause of the rash is unknown.\\
    No causality at all & The patient is in room 302. \\
    \bottomrule
\end{tabularx}
\end{table}
Our main contribution is a systematic benchmark of open-source LLMs for PCD. As shown in Table 2, our evaluation is more comprehensive than prior work. Crucially, we evaluate models on a diverse collection of datasets spanning healthcare, finance, and social science. This multi-domain approach stress-tests generalizability, distinguishing true causal reasoning from simple pattern matching of medical jargon. A model that performs well across these contexts is more likely to be a reliable tool for the complex linguistic environment of healthcare.

\begin{figure*}[htb]
  \centering
  \small
\includegraphics[page=1,width=0.99\textwidth]
  {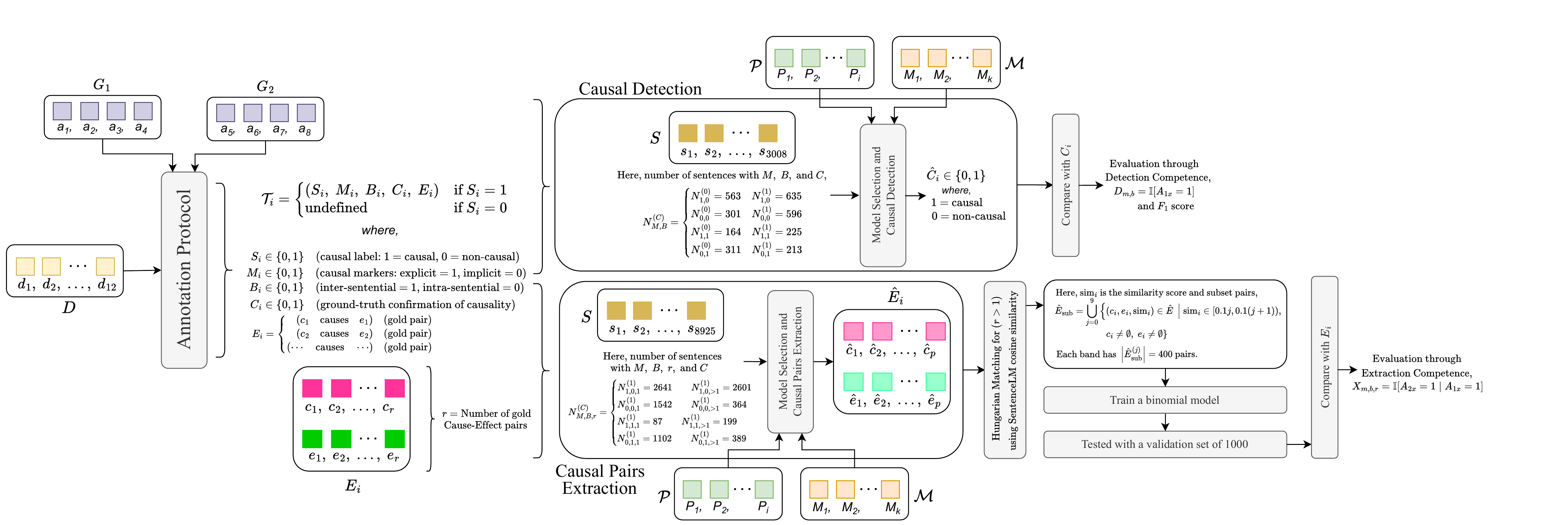}
  Fig. 1: {Overview of the experimental design of PCD pipeline. Here, 12 datasets $\ds$ are taken as the input of the annotation protocol, then 2 groups of annotators $\ann$ produce golden dataset with C, M, B, C, E (described in Study Design Section). Then, subsets of sentences $\sen$ are used with different combinations of prompts $\prompt$ and 13 LLM models $\model$ for \textbf{Causal Detection} and \textbf{Causal Pairs Extraction}. The Causal Detection model gives the output of $\hat C_i$, whereas the Causal Pairs Extraction model gives cause $\cause$ and effect $\effect$ pairs $\hat E_i$ as output. At the end, we analyze the detection and extraction competence to validate our method.}
\label{fig:framework}
\end{figure*}

The key highlights of our study are:
\begin{enumerate}
    \item A comprehensive benchmark of LLMs on  PCD, spanning multiple dataset types (e.g., intrasentential/inter-sentential, explicit/implicit causality markers).
    \item Twelve experimental setups to test robustness across different conditions, thus providing various metrics of LLMs performance in these unique instances.
    \item Assessment of thirteen open-source LLMs.
    \item Detailed analysis of performance on PCD.
\end{enumerate}

We study PCD, first detecting true causality, then extracting cause, effect pairs. Section 2 provides a background on CD and the motivation for this study; Section 3 details our study design, including the 12 datasets, our rigorous annotation protocol, and model selection; Section 4 presents our methodology, detailing the prompting strategies and evaluation metrics.; Section 5 reports results; Section 6 discusses them; Section 7 lists limitations; Section 8 concludes.

\section{Background}
\subsection{Causal Discovery, (CD)}

CD is the systematic process of identifying underlying cause-and-effect relationships from data. One part of CD described by Glymour, Zhang, and Spirtes (2019) \cite{glymour2019review}, involves a causal structure search which delves into analyzing statistical properties of observational data. Eberhardt (2017) \cite{eberhardt2017introduction} framed it as the task of inferring as much as possible about the underlying causal graph from the joint distribution of observed variables. CD is key in distinguishing true causal mechanisms from mere statistical associations, particularly when controlled experiments are infeasible. Its importance can be emphasized in settings where there is a need for formulating valid inferences, guiding effective interventions, and informing decision-making in scientific, medical, and policy domains.

In Judea Pearl’s original formulation of the three-rung ladder of causality\cite{pearl2009causality}, CD was not explicitly identified as a separate category; instead, it was subsumed under level-1 association. However, a recent survey by Jing Ma
(2025) \cite{ma2025causalinferencelargelanguage} positions CD as preceding association, effectively implying it as a level-0 rung in the causality ladder. This reclassification is a reflection of a growing consensus in the literature, particularly in the context of LLMs, where the initial challenge is to establish their capacity for CD before advancing to more complex forms of causal inference. A fundamental aspect of CD is pairwise causal discovery (PCD), which centers on identifying directional relationships between individual pairs of variables, namely, distinguishing the cause from the effect. This identification is a critical first step in evaluating the capability of LLMs for causal reasoning tasks.

 PCD is also fundamental to creating causal graphs. LLMs are becoming a common tool for building these graphs \cite{wan2025largelanguagemodelscausal}, but their ability to correctly identify causal connections from text is a significant challenge. The ideal process would be to apply PCD techniques to factual text to infer the initial connections and their directions. This first draft of the graph could then be improved with more domain knowledge or statistical rules. Causal graphs, by creating a more complete picture of how things are connected rather than just isolated pairs, become an essential tool for deeper analysis, such as causal inference, intervention analysis, and policy evaluation. For this reason, it is crucial that LLMs improve at PCD.

\subsection{Motivation}
\textit{LLMs for Causality research} evaluates the effectiveness of LLMs as tools for performing causal inference tasks \cite{wan2025largelanguagemodelscausal}. Among these, CD stands out as both an essential and particularly challenging problem. Typical evaluation strategies include LLMs having to identify, extract, or predict causal relationships from data or unstructured text, and determining their reliability and accuracy compared to traditional CD methods. However, traditional CD methods rely on carefully engineered statistical algorithms and strong assumptions about data distributions. These limitations often restrict their applicability in complex domains like healthcare, where data is often observational, high-dimensional, and unstructured (e.g., from Electronic Health Records), and controlled interventions are not feasible.

\begin{table*}[t]
\centering
\scriptsize
Table 2: {Datasets used for analysis}
\renewcommand{\arraystretch}{1.1}
% Three fixed-width columns that sum < \textwidth; @{} trims outer padding.
\begin{tabular}{@{}p{0.15\textwidth} p{0.15\textwidth} p{0.20\textwidth} p{0.35\textwidth}@{}}
\toprule
\textbf{Dataset} & \textbf{Focus} & \textbf{Domain} & \textbf{Size \& Key Features} \\
\midrule
CausalNet~\cite{ashwani2024cause}                   & causal reasoning (CR) & General Domain \& News           & 1K scenarios; cause–effect \& CF Q\&A; multi‐domain \\
CausalProbe~\cite{chi2024unveiling}            & news‐based CR & Multi-Domain \& Abstract & 3.5K Q\&A; easy/hard/MC; recent articles \\
SemEval2010Task8~\cite{hendrickx-etal-2010-semeval} & semantic relations         & General Domain \& News & 10.7K sentences; 9 rel. types; cause–effect subset \\
MedCaus~\cite{moghimifar-etal-2020-domain}          & medical causality          & Biomedical / Medical & 15K sentences; explicit/implicit/nested/non‑causal \\
CauseNet~\cite{10.1145/3340531.3412763}               & open‑domain causal KB       & General Domain \& New & 11M relations; 83\% full; 96\% high‑prec. subset \\
FinCausal~\cite{mariko-etal-2020-financial}         & financial causality         & Financial & 29K segments; binary detection; span extraction \\
CausalBench~\cite{wang-2024-causalbench}            & multi‑domain CR & Multi-Domain \& Abstract & $>$ 60K problems; text/math/code \\
CRASS~\cite{frohberg-binder-2022-crass}             & counterfactual reasoning    & Multi-Domain \& Abstract & 274 PCTs; premise–CF tuples \\
Coling22~\cite{schlatt-etal-2022-mining} & health causality     & General Domain \& News)  & 7.8M statements; phrase/sent. level \\
ECI-B~\cite{wang-chen-etal2022MAVENERE}         & event relation extraction    & General Domain \& News & 4.5K docs; 58K causal rels.; temporal \& subevent \\
PubMed~\cite{white2020pubmed}                                & biomedical abstracts     & Biomedical / Medical     & $>$ 30M abstracts; life‑sciences \& clinical \\
COPA~\cite{roemmele2011choice}                      & commonsense CR & Multi-Domain \& Abstract & 1K instances; premise + 2 alternatives \\
\bottomrule
\end{tabular}
\label{tab:datasets}
\end{table*}

LLMs, with their advanced natural language understanding, offer a promising alternative for performing CD tasks directly on unstructured texts like biomedical abstracts, clinical notes, or EHRs. If LLMs can be effectively leveraged for this, there is hope of automating the first rung of causal inference, pairwise CD. This task, which has historically been a bottleneck dependent on manual human annotation for tasks like systematic reviews or drug-side-effect databases, is fundamental to medical knowledge generation. \textcolor{black}{Finetuning LLMs that can ascertain and distinguish causal relationships will accelerate the speed and scale at which health policy makers can make informed decisions from text, such as identifying adverse drug events or validating novel therapeutic targets.} \textcolor{black}{From our review \cite{wan2024large, wu2024causalitylargelanguagemodels}, we have seen that the ability of LLMs to perform CD unsupervised with just prompting has been underexplored and empirically untested, particularly in the instance of PCD.} We therefore aim to benchmark and evaluate the effectiveness of open-source LLMs on PCD tasks across a diverse range of datasets and scenarios, including biomedical contexts. We curated a list of articles in Table 2 which have tested PCD, but none is as exhaustive as our evaluation process. Our work is positioned to be a guide to test the current capabilities of open-source LLMs in PCD causal reasoning and therefore, inform the development of more robust, reliable, and interpretable models for real-world causal inference in high-stakes fields like medicine.

\section{Study Design}

In the previous section, we discussed how existing work covered different aspects of CD including detection and extraction. Building on these findings, our experiments will benchmark open-source LLMs on PCD across health, social science, and physical-process datasets using zero-shot, few-shot, and fine-tuned approaches. Motivated by automating the first rung of Pearl’s causality ladder, we’ll measure directional accuracy, domain robustness, and prompt sensitivity, expecting that fine-tuning yields the largest gains while clever prompting offers modest improvements. 

\subsubsection{Datasets}
{We utilize a diverse collection of 12 datasets across different domains} present in Table 2. For each of the experiments we designed, we curated the distribution of the 12 datasets in the unattached code and data page.

\begin{table*}[t]
\footnotesize
\centering
    Table 3: {Coverage matrix. $\checkmark$ explicit; $\sim$ implied; x none. These literature discusses how each paper has handled a single part of the problem.}
\vspace{3pt} 
\begin{tabular}{|p{8.5cm}|*{12}{>{\centering\arraybackslash}p{1em}|}}
    \hline
    Reference & 1a & 1b & 1c & 1d & 2a & 2b & 2c & 2d & 2e & 2f & 2g & 2h \\
    \hline
    Sun, Chao, and Li 2023 \cite{sun2023event}                                                    & x & x & x & x & $\checkmark$ & x & x & x & x & x & x & x \\ \hline
    Ban et al. 2023 \cite{ban2023query}                                                    & x & x & x & x & $\checkmark$ & x & x & x & $\checkmark$ & x & x & x \\ \hline
    Schrader et al. 2023 \cite{schrader2023boschai}                                            & x & x & x & x & $\checkmark$ & $\checkmark$ & x & x & x & x & x & x \\ \hline
    Kiciman et al. 2023 \cite{kiciman2023causal}                                              & $\sim$ & $\sim$ & x & x & x & x & x & x & x & x & x & x \\ \hline
    Feng et al. 2024 \cite{feng2024pre}                                                    & $\checkmark$ & $\checkmark$ & $\sim$ & $\sim$ & x & x & x & x & x & x & x & x \\ \hline
    Jin et al. 2020; Gopalakrishnan et al. 2023; Haoang et al. 2024;
Hobbhahn, Lieberum, and Seiler 2022 \cite{jin2020inter,gopalakrishnan2023text,chi2024unveiling,hobbhahn2022investigating}
    & x & x & x & x & $\sim$ & $\sim$ & $\sim$ & $\sim$ & x & x & x & x \\ \hline
   Ahne et al. 2022 \cite{ahne2022extraction}                                             & $\sim$ & $\sim$ & x & x & $\sim$ & $\sim$ & x & x & x & x & x & x \\ \hline
   Liu et al. 2023 \cite{liu2023event}                                                   & x & x & x & x & $\checkmark$ & $\checkmark$ & x & x & $\checkmark$ & $\checkmark$ & x & x \\ \hline
    Gopalakrishnan, Garbayo, and Zadrozny 2024 \cite{gopalakrishnan2024causality}                                   & x & x & x & x & $\sim$ & $\sim$ & x & x & $\sim$ & $\sim$ & x & x \\ \hline
    Li et al. 2021; Chen and Mao 2024 \cite{li2021causality,chen2024explicit}                               & x & x & x & x & $\checkmark$ & $\checkmark$ & x & x & $\checkmark$ & $\checkmark$ & x & x \\ \hline
    Takayanagi et al. 2024 \cite{takayanagi2024chatgpt}                                          & x & x & x & x & $\checkmark$ & $\checkmark$ & $\sim$ & $\sim$ & $\checkmark$ & $\checkmark$ & $\sim$ & $\sim$ \\ \hline
    Wang et al. 2024 \cite{wang2024document}                                               & x & x & x & x & $\checkmark$ & $\checkmark$ & $\checkmark$ & $\checkmark$ & x & x & x & x \\ \hline
    Our Evaluation                                                        & $\checkmark$ & $\checkmark$ & $\checkmark$ & $\checkmark$ & $\checkmark$ & $\checkmark$ & $\checkmark$ & $\checkmark$ & $\checkmark$ & $\checkmark$ & $\checkmark$ & $\checkmark$ \\
    \hline
  \end{tabular}
  \label{tab:coverage}
\end{table*}

\subsubsection{Experiment Tasks}

Prior research in PCD often aggregates results, without systematically evaluating the abilities of large language models (LLMs) across different types of causal statements. To address this, we design twelve experiments, spanning both detection and extraction tasks, the two reasoning steps in PCD. This is organized by the presence of explicit causal markers, sentence boundaries, and causal complexity. Each data point is:
\begin{equation}
(S =1 : M \in \{0,1\},\; B \in \{0,1\},\; C \in \{0,1\},\; E \subseteq \mathcal{E}(S)),
\end{equation}
%\sun{use ; instead of , -- somehow it seems that both S and M are in \{0,1\}}

%\sun{also see if you can add examples of each case - marked, unmarked etc. If there's no space, add it in the appendix}\sun{can we call it a sentence or say text span, such as a sentence?}

where $S = 1$ represents causal sentence(s), $C$ is the ground-truth causality label, $E$ is the set of gold cause–effect (C–E) pairs, and our three main axes of difficulty are:

\begin{itemize}[leftmargin=*]
    \item \textbf{Marker Presence ($M$):} Explicit ($M=1$, containing clear lexical cues of ``causes'' and ``results'') vs. Implicit ($M=0$, unmarked).
    \item \textbf{Textual Scope ($B$):} Intra-sentential ($B=0$, within one sentence) vs. Inter-sentential ($B=1$, spanning multiple sentences).
    \item \textbf{Cardinality ($r$):} The number of gold C-E pairs. This is set to a Single Pair ($r=1$) or Multiple Pairs ($r>1$).
\end{itemize}

We partition the 12 experiments based on these variables as follows:

\subsection*{a. Causal Detection (Tasks 1a--1d):}
The model determines if a causal relation exists in $S$ (i.e., if $C=1$).

\begin{itemize}[leftmargin=*,labelwidth=2em,align=left]
    \item[\textbf{1a:}] $M=1, B=0$ (Explicit, Intra-sentential)
    \item[\textbf{1b:}] $M=0, B=0$ (Implicit, Intra-sentential)
    \item[\textbf{1c:}] $M=1, B=1$ (Explicit, Inter-sentential)
    \item[\textbf{1d:}] $M=0, B=1$ (Implicit, Inter-sentential)
\end{itemize}

\subsection*{b. Causal Extraction (Tasks 2a--2h):}
Given $S$ with $C=1$, the model extracts all C-E pairs.

\begin{itemize}[leftmargin=*,labelwidth=2em,align=left]
    \item[\textbf{2a:}] $M=1, B=0, r=1$ (Explicit, Intra, Single Pair)
    \item[\textbf{2b:}] $M=0, B=0, r=1$ (Implicit, Intra, Single Pair)
    \item[\textbf{2c:}] $M=1, B=1, r=1$ (Explicit, Inter, Single Pair)
    \item[\textbf{2d:}] $M=0, B=1, r=1$ (Implicit, Inter, Single Pair)
    \item[\textbf{2e:}] $M=1, B=0, r>1$ (Explicit, Intra, Multiple Pairs)
    \item[\textbf{2f:}] $M=0, B=0, r>1$ (Implicit, Intra, Multiple Pairs)
    \item[\textbf{2g:}] $M=1, B=1, r>1$ (Explicit, Inter, Multiple Pairs)
    \item[\textbf{2h:}] $M=0, B=1, r>1$ (Implicit, Inter, Multiple Pairs)
\end{itemize}

An example of each experiment class in Experiment 1 and 2 will be added to the Code and Data page, which is not attached to this document because of the publication's requirements.
%\sun{add a table with one example from each category} -done \\

In the detection tasks, for Experiment 1a and 1c, the false class contains causal markers which bear no relationship to causality at all. For instance, the statement ``the false class contains causal markers", has no attribution to true causality, therefore, this would stump the model especially if the model uses cue words to define causality rather than genuine causal reasoning. Experiment 1b and 1d false classes are sentences that have no causality, or are implied. We also defined causal claims as not having causality. For the extraction tasks, all examples are causal and we test how well we can extract the entity pairs from the text using LLMs.

\subsubsection{Annotation Protocol}

We recruited eight annotators based on a screening test (score $\ge83\%$). They received a detailed guideline and worked in two groups of four. To ensure consistency and reliability, all annotators adhered to a strict set of principles during the labeling process:
\begin{itemize}[leftmargin=*]
\item {Factual Assertions Only:} A causal link was labeled only if the text stated it as a fact. Reported claims (e.g., ``Scientists believe X causes Y''), hypotheticals, or negated effects were treated as non-causal.
\item {Unambiguous Spans:} Both the cause and effect spans had to be clearly and unambiguously identifiable in the text. If either was vague or relied on an abstract reference (e.g., ``It causes..."), the instance was labeled non-causal.
\item {Minimal Granularity:} Annotators were instructed to select the smallest contiguous phrase that fully expressed the cause and effect, while explicitly excluding causal cue words in the cause-effect pairs (e.g., ``because," ``leading to").
\item {Decomposition of Relations:} Sentences containing multiple causes or effects were decomposed into distinct pairwise relations. For instance, ``A causes B and C" was annotated as two separate pairs: (Cause: A, Effect: B) and (Cause: A, Effect: C).
\item {Default to Non-Causal:} In any case of ambiguity or uncertainty, the default instruction was to label the instance as non-causal.
\end{itemize}
Disagreements were adjudicated by a senior reviewer and the authors; only instances with 100\% agreement were retained for the final dataset. Cause and effect spans were scored using Hungarian matching and a calibrated SentenceLM-based similarity scheme.

\subsubsection{Model Choices}
The motivation behind the \textcolor{black}{choice of these} models was to cover various sizes (3B--70B) and architectures (dense, MoE), with both base and instruction variants, which ensures a robust comparison, offering insights on cost and accuracy trade-offs. We use the following models in Table 4.

\begin{table}[ht]
\centering
\footnotesize
     Table 4: Shorthand of Selected Models used in this narrative.

\vspace{3pt} 
%\scriptsize

%\renewcommand{\arraystretch}{1.1}
\setlength{\tabcolsep}{4pt}
% Left: table occupying roughly 1.5x a single column (here ~0.65\textwidth), right: text block (~0.33\textwidth)
 %\vspace{3pt}
 \resizebox{\columnwidth}{!}{%
  \begin{tabular}{ll}
    \toprule
    \textbf{Shorthand} & \textbf{Full Name} \\
    \midrule
    DS-R1-0528-Q3-8B    & DeepSeek-R1-0528-Qwen3-8B        \\
    DS-R1D-L70B         & DeepSeek-R1-Distill-Llama-70B      \\
    DS-R1D-Q32B         & DeepSeek-R1-Distill-Qwen-32B     \\
    L3-8B               & Llama-3-8B       \\
    L3.1-8B-I           & Llama-3.1-8B-Instruct       \\
    L3.2-3B             & Llama-3.2-3B      \\
    L3.2-3B-I           & Llama-3.2-3B-Instruct     \\
    MetaL-3.1-8B        & Meta Llama-3.1-8B     \\
    MetaL-3.3-70B-I     & Meta Llama-3.3-70B-Instruct    \\
    Mistral-7B-I-0.3    & Mistral-7B-Instruct-v0.3    \\
    Mixtral-8x7B-I-0.1  & Mixtral-8x7B-Instruct-v0.1    \\
    Qwen2.5-7B-I        & Qwen2.5-7B-Instruct          \\
    Qwen2.5-C-32B-I & Qwen2.5-Coder-32B-Instruct    \\
    \bottomrule
  \end{tabular}
 }
  \label{tab:model-names}
\end{table}

Moreover, since these models are well-supported open models (Llama 3.x, Qwen 2.5, Mistral/Mixtral, DeepSeek\mbox{-}R1 distills), it would be easier to obtain reproducible benchmarks and they would be relatively straightforward in real-world  deployment. Furthermore, our models set includes reasoning-tuned (DeepSeek\mbox{-}R1, distills) and code-tuned (Qwen Coder) models to evaluate reasoning, instruction following, and coding---our primary workload mix.

\begin{table*}[ht]
  \centering
  \scriptsize
  \setlength{\tabcolsep}{3pt}
  Table 5: {Model-wise Distributions of $D$ and $X$ Metrics with MOE ($< 5\%)$ in all cases}
  \label{tab:results_metrics}
  \begin{tabular}{l*{14}{r}}
    \toprule
    \textbf{Model}
      & $D_{1,0}$ & $D_{0,0}$ & $D_{1,1}$ & $D_{0,1}$
      & $X_{1,0,1}$ & $X_{0,0,1}$ & $X_{1,1,1}$ & $X_{0,1,1}$
      & $X_{1,0,>1}$ & $X_{0,0,>1}$ & $X_{1,1,>1}$ & $X_{0,1,>1}$ &  $C_{\mathrm{detect}}$
      & $C_{\mathrm{extract}}$ \\
    \midrule
    DS-R1-0528-Q3-8B    & 3.42\% & 5.91\% & 4.88\% & 2.29\% & 13.86\% & 12.61\% & 9.76\%  & 38.05\% & 3.27\%  & 7.12\%  & 2.50\%  & 5.66\%  & 4.13\%  & 11.60\%   \\
    DS-R1D-L70B         & 47.66\%& \textbf{66.00\%}& \textbf{50.64\%}& \textbf{33.97\%}& 5.03\%  & 5.39\%  & 1.18\%  & 7.99\%  & 0.08\%  & 0.00\%  & 0.00\%  & 0.00\%  & \textbf{49.57\%} & 2.46\%    \\
    {DS-R1D-Q32B}& \textbf{49.37\%}& 63.21\%& 47.30\%& 33.40\%& 30.82\% & 28.76\% & 29.73\% & 59.95\% & 13.26\% & 12.09\% & 22.77\% & 41.90\% & 48.32\% & 29.91\%  \\
    L3-8B               & 9.27\% & 15.16\%& 13.11\%& 5.15\% & 26.13\% & 20.00\% & 57.89\% & 47.59\% & 9.39\%  & 6.85\%  & 13.43\% & 38.30\% & 10.67\% & 27.45\%   \\
    L3.1-8B-I           & 3.01\% & 2.45\% & 2.57\% & 2.10\% & 28.97\% & 25.15\% & 42.67\% & 36.78\% & 13.59\% & 16.39\% & 34.95\% & 56.56\% & 2.53\%  & 31.88\%   \\
    L3.2-3B             & 0.50\% & 1.45\% & 1.80\% & 0.95\% & 33.77\% & 15.73\% & 23.68\% & 34.60\% & 5.47\%  & 2.47\%  & 7.43\%  & 21.85\% & 1.18\%  & 18.13\%   \\
    L3.2-3B-I           & 33.72\%& 48.72\%& 33.93\%& 25.19\%& 21.28\% & 10.79\% & 16.25\% & 24.36\% & 8.33\%  & 4.93\%  & 5.97\%  & 22.11\% & 35.39\% & 14.25\%  \\
    MetaL-3.1-8B        & 22.29\%& 31.66\%& 24.94\%& 20.61\%& 30.30\% & 23.20\% & 63.01\% & 59.71\% & 9.93\%  & 8.52\%  & 16.18\% & 43.44\% & 24.87\% & 31.79\%   \\
    MetaL-3.3-70B-I     & 12.27\%& 22.07\%& 18.25\%& 9.92\% & 38.39\% & 25.02\% & 42.11\% & {36.26\%}& 18.25\% & 17.76\% & \textbf{40.49\%}& 65.04\% & 15.63\% & 35.41\%   \\
    Mistral-7B-I-0.3    & 41.15\%& 60.20\%& 41.39\%& 29.39\%& 35.76\% & 37.73\% & 34.62\% & 63.85\% & 12.50\% & 29.95\% & 23.90\% & 54.76\% & 43.03\% & 36.63\%  \\
    Mixtral-8x7B-I-0.1  & 38.15\%& 46.93\%& 31.88\%& 22.14\%& 28.94\% & {41.68\%}& {60.81\%}& \textbf{68.06}\% & 1.56\%  & 3.29\%  & 3.02\%  & 6.17\%  & 34.77\% & 26.69\%  \\
    Qwen2.5-7B-I        & 8.85\% & 19.06\%& 8.23\% & 11.45\%& 35.34\% & 46.97\% & 56.76\% & 45.92\% & 13.27\% & {28.42\%}& 35.61\% & \textbf{69.92\%} & 11.90\% & 41.53\%  \\
    Qwen2.5-C-32B-I & 25.54\%& 43.03\%& 23.14\%& 15.46\%& \textbf{42.62\%}& \textbf{50.72\%}& \textbf{71.23\%}& 59.64\%& \textbf{23.93\%}& \textbf{36.89\%}& 33.82\% & 58.10\% & 26.79\% & \textbf{47.12\%}  \\
    \bottomrule
  \end{tabular}
\end{table*}

\subsubsection{Evaluation Definitions}
\begin{itemize}
    \item \emph{Detection competence:} $D_{m,b} = \mathbb{I}[A_{1x}=1]$ This is the probability of correct causal detection for marker flag $m$ and boundary flag $b$.
    \item \emph{Extraction competence:} $X_{m,b,r} = \mathbb{I}[A_{2x}=1\,|\,A_{1x}=1]$ This is the probability of extracting all gold C–E pairs given successful detection, for marker $m$, boundary $b$, and $r$ pairs.
    \item Overall task success factors as
$
\mathbb{I}[A_{2x}=1] = D_{m,b} \, X_{m,b,r}
$

%\subsubsection{Experimental Tasks}
%Our evaluation framework is structured around 12 distinct experimental setups, categorized into two primary objectives: \textbf{Causal Detection } and \textbf{Causal Extraction}. This design allows us to systematically measure performance across different linguistic and structural challenges by comparing the results of the models annotations to a human professional.

    % \item Conditional extraction probability after detection:
    % We can calculate the probability that, once the model has correctly detected causality, it will succeed on either the single or multiple C–E extraction sub-task. It shows how extraction competence splits between these two cases.
    % $$
    % \mathbb{I}[A_{2a} \vee A_{2e}\mid A_{1a}] = \frac{1}{2} (X_{1,0,1} + X_{1,0,>1})
    % $$
    % \item Macro-competence: For each model, we average the detection and extraction competencies across all possible  settings, giving us a single score.

    % $$
    % C_{\text{detect}} = \frac{1}{4}\sum_{m,b} D_{m,b}; \text{ } 
    % C_{\text{extract}} = \frac{1}{8}\sum_{m,b}\sum_{r \in \{1,>1\}} X_{m,b,r}
    % $$
    % \item Headline probability of success: Finally, we look into each models probability to answer all detection and at least one extraction sub-task in every condition, combining the probabilities of success across all experimental settings.
    % $$
    % \mathbb{I}[\text{headline}] = \mathbb{I}od_{m,b} D_{m,b} \mathbb{I}od_{m,b}( \frac {l}{l+h}X_{m,b,1} + \frac{h}{l+h}X_{m,b,>1})
    % $$
\end{itemize}

% \textbf{Evaluation Metrics.}
% For detection on any slice $\mathcal{D}$:
% $$
% \text{TPR} = \mathbb{I}[Y=1 \mid C=1] \text{, FPR} = \mathbb{I}[Y=1 \mid C=0]
% $$
% $$
% \mathcal{L}_{\text{CE}} = -\frac{1}{|\mathcal{D}|} \sum_{i\in\mathcal{D}} [C_i \log \hat{p}_i + (1-C_i)\log(1-\hat{p}_i)]
% $$

% For C–E extraction on $\mathcal{D}$:
% $$
% \text{Precision} = \mathbb{E}[\text{prec}] \text{, Recall} = \mathbb{E}[\text{rec}]
% $$
% $$
% \text{F1} = \frac{2\,\text{Precision}\,\text{Recall}}{\text{Precision}+\text{Recall}}
% $$

% \textbf{Parameter Estimation.}
% Given observed pass rates $\widehat{p}_{1x}, \widehat{p}_{2x}$:
% $$
% \hat{D}_{m,b} = \widehat{p}_{1x} \text{, } \hat{X}_{m,b,r} = \frac{\widehat{p}_{2x}}{\widehat{p}_{1x}}
% $$

\subsubsection{Prompting Strategies}
To systematically evaluate model performance, we employed a range of prompting strategies, progressing from simple zero-shot instructions to complex, multi-step reasoning frameworks. This approach allows us to measure both baseline instruction-following and more advanced reasoning capabilities. The strategies were selectively applied to the detection and extraction tasks based on their suitability.

\subsection*{a. Instruction-only (Zero-shot)}
This baseline strategy provides a direct, zero-shot instruction to the model. The prompt defines the task, specifies the strict output format, and provides the input text. No examples are given. This tests the model's ``out-of-the-box" ability to understand and execute the CD tasks. A representative template for causal detection is shown in Listing \ref{lst:zero-shot}.

\begin{listing}[tb]
\scriptsize
\begin{lstlisting}[language={}, captionpos=b, caption={Zero-shot instruction prompt for Causal Detection.}, label={lst:zero-shot}]
You are a causal reasoning expert. Your task is to identify if the text contains a factual causal relationship.
- A ``Causal" relationship is one where a cause is stated as a fact to produce an effect.
- A ``Non-Causal" relationship includes correlations, reported claims (``studies show"), or hypotheticals.

Analyze the following text.
Text: {INPUT_TEXT}

Does this text contain a causal relationship? Answer only ``Causal" or ``Non-Causal".

Answer:
\end{lstlisting}
\end{listing}

\subsection*{b. Few-shot In-Context Learning (FICL)}
For FICL, we prepended $k=3$ canonical examples of the task to the instruction-only prompt. These examples (text and the corresponding label or JSON output) were selected to demonstrate the task's nuances, such as the distinction between causation and correlation. This tests the model's ability to learn the task from in-context examples.

\subsection*{c. Chain-of-Thought (CoT) and Hybrid CoT+FICL}
To elicit more complex reasoning, we tested two variants of Chain-of-Thought (CoT).

\begin{itemize}[leftmargin=*]
    \item {CoT-only:} We appended the phrase ``Let's think step by step:" to the zero-shot instruction, prompting the model to generate a rationale before its final answer.
    \item {Hybrid CoT+FICL:} We combined FICL and CoT. The $k=3$ in-context examples were augmented to include a brief rationale (e.g., ``Rationale: The sentence explicitly states 'was caused by', indicating a factual causal link. The cause is 'the fall' and the effect is 'the fracture'.") before the final label. This hybrid approach, applied to detection tasks, proved to be the most effective, as noted in our abstract. A template for extraction is shown in Listing \ref{lst:cot-ficl-extraction}.
\end{itemize}

\begin{listing}[tb]
\begin{lstlisting}[language={}, captionpos=b, caption={Hybrid CoT+FICL prompt for Causal Extraction (2a).}, label={lst:cot-ficl-extraction}]
You are a causal reasoning expert. Your task is to extract the exact cause and effect spans from the text.
...[Detailed instructions and FICL examples with rationales]...

Text: {INPUT_TEXT}

Rationale: Let's think step by step to identify the cause and effect.
1. [Model generates reasoning step 1]
2. [Model generates reasoning step 2]
...
Output:
\end{lstlisting}
\end{listing}

\subsection*{d. Advanced Extraction Prompts}
For the more complex extraction tasks (2a-2h), we also evaluated two decompositional methods mentioned in our abstract.
\begin{itemize}[leftmargin=*]
    \item {Least-to-Most (LtM):} This strategy decomposes the task into a sequence of simpler sub-problems. We first prompted the model to identify and extract only the cause span. Then, using the text and the identified cause span as input, we prompted the model to identify and extract the corresponding effect span.
    \item {ReAct:} This framework prompts the model to generate alternating reasoning traces and task-specific actions (e.g., `Thought`, `Action`, `Observation`). This was used to navigate the most complex multi-pair, inter-sentential tasks (2g, 2h), though it showed high variance in output structure.
\end{itemize}

\section{Methodology}
% Ensure your preamble includes \usepackage{booktabs,array}

% Our goal defined in the study design is to obtain reliable estimates of the detection and extraction competences 
% $$
% D_{m,b} \quad\text{and}\quad X_{m,b,r}
% $$

\subsubsection{Dataset Curation}
Most public corpora either did not properly tag the causal markers or lack inter-sentential labels. We first ran lightweight heuristics for each dataset to look for datasets with both intersentential and intrasentential spans \(S\). We then constructed disjoint sets for each slice with a selected seed value which was for prompt tuning validation. We froze the selected prompt and executed a single pass on the test split per model. All data and prompt files are provided in the Technical Appendix.

\subsubsection{Annotation Protocol}
Eight annotators were recruited via a screened test   (score $\ge83\%$).  
They received a detailed guideline and worked in two groups, each group containing 4 individuals. Disagreements were adjudicated by a senior reviewer (a recruited expert), and the authors, and only rows with 100\% agreement were selected for the final distillation. The Cohen's kappa between the annotators before retaining only the final agreement was an average of $\kappa \ge 0.758$, across the experiments.  All details of the annotation are present in the Technical Appendix which contains the Cohen's kappa scores for each experiment, the annotation policies used for the dataset, the recruitment form and the questions used to hire the annotators.
%\sun{include all details of hiring annotators, recruiting criteria, testing the quality of their response, include samples}

\subsubsection{Prompt Engineering}
We benchmark four prompting styles for Experiment 1: (instruction‑only, few‑shot in‑context learning (FICL), chain‑of‑thought (CoT), and hybrid CoT+FICL) and five prompting styles for Experiment 2: (general instruction‑only (GIP), few‑shot in‑context learning (FICL), chain‑of‑thought (CoT), Least-to-Most (L2M) and React (RCT)). We report all the reproducibility metrics like seed=4000, temperature=0.7, batch size=64, top-k=0.7, timings and the config-file in the Technical Appendix.

\subsubsection{Scoring of the Models}
For items with multiple cause–effect pairs, we perform Hungarian 1:1 matching between model outputs and gold pairs. We first compute a similarity matrix using SentenceLM-based cosine scores for every predicted–gold pair. We then iteratively select the highest-scoring unmatched pair, removing its row and column until either predictions or gold pairs are exhausted; unmatched predictions count as false positives and unmatched gold pairs as false negatives.

\begin{table}[ht]
\centering
\scriptsize
  \setlength{\tabcolsep}{3pt}
  Table 6: {Probabilistic Score Map for SentenceLM Bands}
\label{tab:score_map}
\begin{tabular}{l*{14}{r}}
\toprule
\textbf{Cosine Band} ($\ge$) & 0.9 & 0.8 & 0.7 & 0.6 & 0.5 & 0.4 & 0.3 & 0.2 & 0.1 & 0.0 \\
\midrule
\textbf{Prob. Score} & 0.99 & 0.72 & 0.62 & 0.59 & 0.33 & 0.26 & 0.13 & 0.06 & 0.03 & 0.1 \\
\bottomrule
\end{tabular}
\end{table}

\begin{figure*}[htb]
  \centering
  \small
\includegraphics[page=1,width=0.99\textwidth]
  {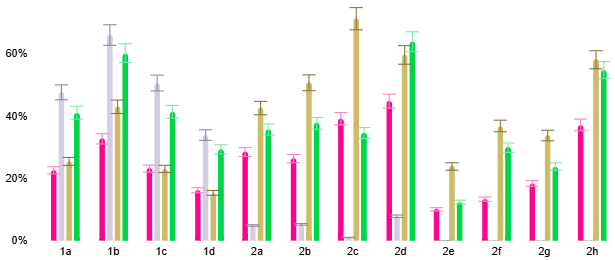}
  Fig. 3: {Performance of the top 3 performing models across the 12 causal detection and extraction tasks. The bar chart shows the average F1-scores for models DS-R1D-L70B $\ann$, Qwen2.5-C-32B-I $\model$, and Mistral-7B-I-0.3 $\effect$, benchmarked against an overall average $\cause$. Each scenario (1a-2h) represents a unique combination of causality type, textual scope, and cardinality (as detailed in Section 3 and 4).} 
    \label{fig:perform_graph}
\end{figure*}

Scoring for Exp.1 (1a-1d) was binary decision, so we evaluated it with a simple Python script.
Exp. 2 (2a-2h) involved 8{,}925 instances evaluated across 13 models and 5 prompting styles, yielding 580{,}125 generations. Exact-string matching is too strict for span outputs, so we adopt approximate matching based on SentenceLM cosine similarity coupled with a binomial mapping. Concretely, we bin the SentenceLM cosine scores for both \emph{cause} and \emph{effect} into 10 bands ($\geq 0 - 0.1$, $\geq 0.1 - 0.2$ ... $\geq 0.9 - 1.0$). For calibration, annotators labeled 400 samples per band (4{,}000 total, seed=4000), and we use the per-band means to derive a probabilistic score map that can be seen in Table 6.

 \normalsize
Using this map, we train a small binomial model that ingests the input plus the predicted pair and outputs a score consistent with the calibrated bands. On a 1{,}000-item human-checked test set (only where the model produced a valid output), the approximation error was 3.28\% (96.72\% accuracy) at the 5\% significance level. At corpus scale, the error dropped to 1.62\%, largely due to a higher proportion of clearly bad responses that are straightforward to score.

\section{Results}

\subsubsection{Detection and Extraction Competence}
From Table 5, we get an idea about how the various models performed through the detection ($D$) and extraction ($X$) tasks in the experiment. While DS-R1D-Q32B shows a higher value for explicit intra-sentential ($D_{1,0}$), DS-R1D-L70B model has a better performance as considered in the other $D$ tasks. Models generally performed better on explicit, single-sentence relations (Task 1a). However, as illustrated in Figure 3, performance dropped significantly when dealing with implicit relations within a single sentence (Task 1b). This is a challenge highly relevant to clinical text analysis, as Figure 4 shows the biomedical/medical datasets comprised a notable portion of this specific test set. This suggest that even within a single sentence, models struggle with the nuanced, marker-less language characteristic of medical notes. 

Figure 3 provides a detailed look at the average F1-scores and those  of the top 3 performing models across all 12 experimental setups (1a-2h). The performance varies dramatically by task. It demonstrates that Qwen2.5-C-32B-I leads on the extraction ($X$) side, coming first in five of eight tasks. The model's strength in explicit, intra-sentential extraction (Task 2a) is particularly noteworthy. Mixtral-8x7B-I-0.1 and Qwen2.5-7B-I are also notably stronger in $X_{0,1,1}$ and $X_{0,1,>1}$ respectively while MetaL-3.3-70B-I stands out with the highest $X_{1,1,>1}$ and a strong $X_{0,1,>1}$.

\begin{figure}[H]
  \centering
  \small
  \label{fig:result}
  \includegraphics[page=1,width=0.5\textwidth]
  {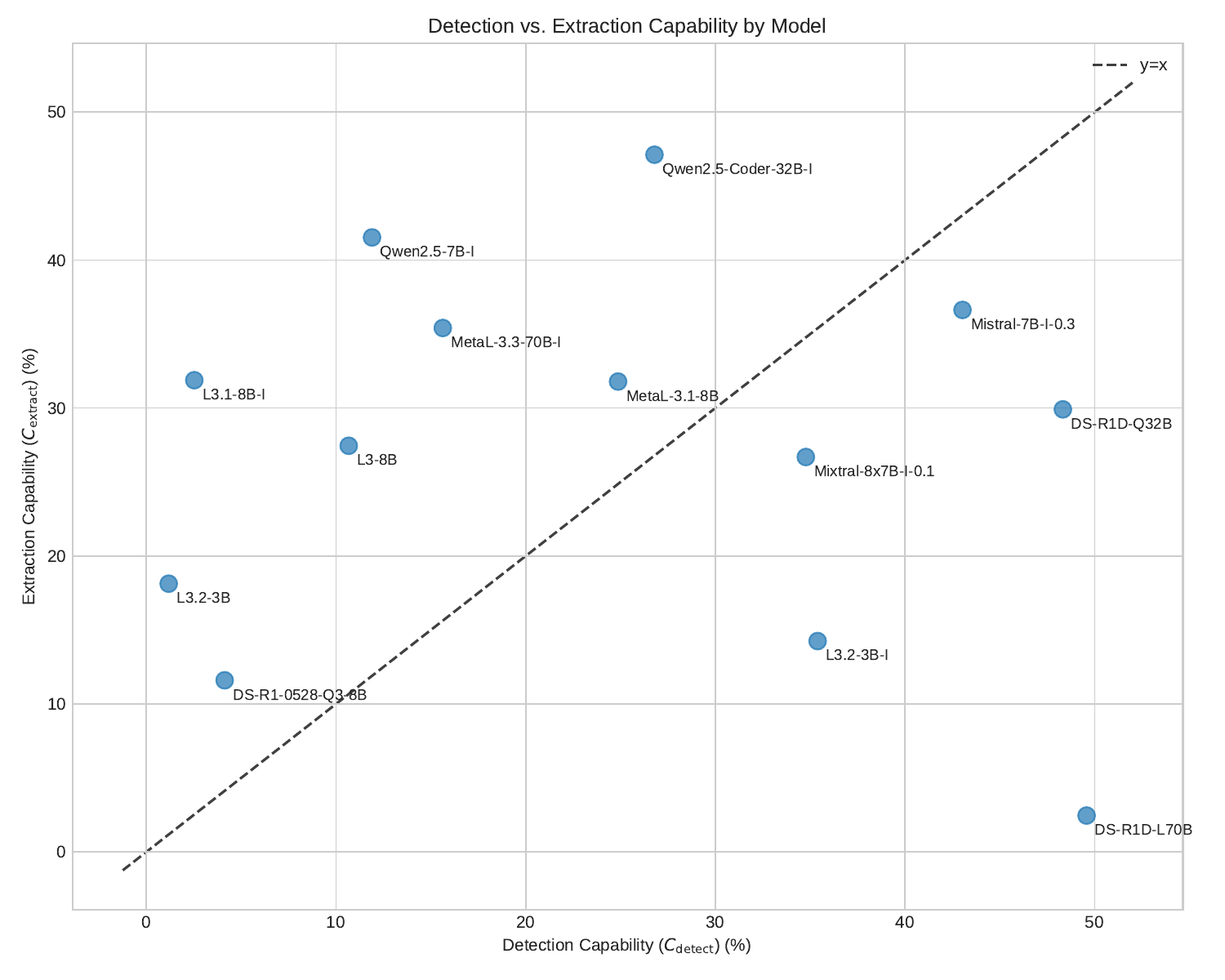}
  Fig. 2: {Trade off between Causal Detection and Causal Extraction in Pairwise Causal Discovery}
\end{figure}

The deepseek models excelled in detection $D$ tasks, but their results in the extraction $X$ tasks were deficient as can be seen in Figure 2. This is not the case in other models like Qwen2.5-C-32B-I, which are more balanced or, in Qwen's case, occupy the top-left quadrant (high extraction, lower detection). This indicates that the process associated with well known causality identification requires different skills than the one of accurate retrieval of cause effect pairs. This could largely be due to the reasoning window of deepseek models. Typically, due to deepseek reasoning taking extra tokens (even when explicitly prompted not to), it shuts off before giving a correct answer, or it forgets to give an answer resulting in missing values. For deepseek models, missing values were the highest in our error analysis, as it recorded approximately 60\%  for this error band. Also, the deepseek models led the detection $D$ tasks since they required straightforward binary answers, so the reasoning did not affect the final answer due to token constraints. 

Each experiment was repeated based on the number of prompts to ensure reproducibility. For the detection case, the Friedman test yielded $\chi^2 = 157.018$, $p < 0.0001$; for the extraction case, $\chi^2 = 229.624$, $p < 0.0001$. We also conducted paired $t$-tests for each experiment. All statistical tests are present in the Appendix.

\subsubsection{End to End Causal Discovery}
\textcolor{black}{Based on Table 5, it can be seen that no individual model stands out across the board. DS-R1D-Q32B and Mistral-7B-I-0.3 are the models with special strengths since they maximize the probabilities on the experiments. DS-R1D-Q32B leads in four tasks ($\mathbb{I}[A2a=1]$, $\mathbb{I}[A2d=1]$, $\mathbb{I}[A2e=1]$ and $\mathbb{I}[A2g=1]$), while Mistral-7B-I-0.3 leads in another three ($\mathbb{I}[A2b=1]$, $\mathbb{I}[A2f=1]$, and $\mathbb{I}[A2h=1]$). As visualized in Figure 2, a clear performance gap is evident when comparing smaller, non-instruction-tuned models (e.g., DS-R1-0528-Q3-8B and L3.2-3B), which perform near zero, to their instruction-tuned counterparts. L3.2-3B-I is the instruction tuned version of the L3.2-3B model, for example, and we can see the improvement in its position along the y-axis in Figure 2.  This indicates how the training strategy followed in instruction tuning affects the CD competence to a great degree.}

\subsubsection{Overall Detection and Extraction Capabilities}
An overview of the model performance can be found in Figure 2, where the columns $C_{detect}$ and $C_{extract}$ from Table 5 are plotted on the x-axis and y-axis, respectively. The DS-R1D-L70B model demonstrates the highest average detection performance ($C_{detect}$) with a score of 49.57\%, indicating its strong overall capability in identifying causal relationships. In contrast, the Qwen models are the top performers in terms of average extraction performance ($C_{extract}$), with Qwen2.5-C-32B-I achieving the highest score of 47.12\%, making it the topmost point on the plot.

Looking at Figure 3, it also leads on implicit tasks 2b and 2f, and 2h. Mistral-7B-I-0.3 and Mixtral-8x7B-I-0.1 also record relatively significant values in these tasks. Mistral-7B-I-0.3 leads  the implicit tasks $\mathbb{I}[A_{2b}\vee A_{2f}\mid A_{1b}]$, $\mathbb{I}[A_{2d}\vee A_{2h}\mid A_{1d}]$, and headline prediction  

\begin{equation}
\mathbb{I}[\text{headline}] = \prod_{m,b} D_{m,b} \prod_{m,b}(X_{m,b,1} + X_{m,b,>1}),
\end{equation}
suggesting it can interpret detection to extraction more effectively as seen in Fig. 2. 

Smaller and non-instruction-tuned models (e.g., DS-R1-0528-Q3-8B and L3.2-3B) perform near zero, while instruction tuning (L3.2-3B-I) produces a massive improvement. MetaL-3.1-8B delivers moderate accuracy, particularly on 2c and 2d, whereas MetaL-3.3-70B-I does not consistently outperform it across all experimental tasks, suggesting sensitivity to its configuration. Qwen2.5-C-32B-I and Mistral-7B-I-0.3 both show multi-target competence; Mistral does not lead on any tasks but it performs close to the peak across all tasks.

\begin{figure}[htb]
  \centering
  \small
  \label{fig:result}
  \includegraphics[page=1,width=0.5\textwidth]
  {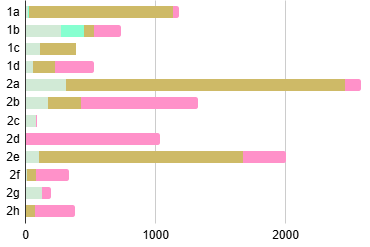}
  Fig. 4: {Domain distribution for each experimental setup (1a-2h). The bars show sample counts per domain: Biomedical/Medical $\prompt$, Financial $\effect$, General Domain \& News $\sen$, and Multi-Domain / Abstract $\cause$. It is important to note that the General Domain \& News and Multi-Domain / Abstract categories also contain healthcare-related text.}
\end{figure}

Crucially, performance on inter-sentential (multi-sentence) extraction offers a direct proxy for a model's ability to process dense biomedical literature. As shown in Figure 4, the test sets for extracting a single explicit pair (2c) and multiple explicit pairs (2g) across sentences were overwhelmingly sourced from the biomedical/medical domain. The strong performance of models like Qwen2.5-C-32B-I and Mixtral-8x7B-I-0.1 in these tasks suggests a nascent capability for navigating the structured language of scientific papers, a key requirement for applications in automated literature review and evidence synthesis.

While our LLM benchmarks yield modest overall scores, these figures are transparently low, reflecting the inherent challenges of unsupervised PCD in diverse, noisy texts. To contextualize, we compare against human baselines from our annotation process: annotators achieved 95\% true positive rate on detection tasks with kappa $\geq$ 0.758 agreement, far outpacing LLMs' average 30-40\%. This gap highlights LLMs' limitations in nuanced reasoning but also their potential as scalable aids; for instance, even partial automation could accelerate biomedical text mining, where humans might take hours per document. Future iterations could incorporate hybrid human-LLM workflows to close this divide, blending model speed with annotator precision.
\section{Analysis and Discussion}
Our analysis reveals that the difficulty of PCD for LLMs is highly dependent on the linguistic structure of the text, a finding with direct implications for their application in healthcare.
\subsubsection{Statistical Test}
We observed several interesting behavioral patterns among models after running 78 t-combination tests on the overall output ($\mathbb{I}[A_{2x}=1]$) for PCD. DS-R1-0528-Q3-8B is demonstrably different from all other models, with no overlapping behavior. Qwen2.5-Coder-32B-Instruct and DeepSeek-R1-Distill-Qwen-32B are statistically indistinguishable from each other ($p \approx 0.45$), yet differ significantly from many mainstream models such as Llama-3.1-8B, Llama-3.2-3B, and Qwen2.5-7B-Instruct. This would be because the distill model is built on the parent model, but we do not always observe this trend. Notably, Qwen2.5-Coder-32B-Instruct shows no significant difference from Mixtral-8x7B-Instruct ($p \approx 0.086$), whereas DeepSeek-R1-Distill-Qwen-32B and Mixtral remain distinct ($p \approx 0.00012$).

We also find a coherent cluster of models with no significant difference which includes Mistral-7B-Instruct, Qwen2.5-7B-Instruct, Meta-Llama-3-8B, Llama-3.3-70B-Instruct, and DeepSeek-R1-Distill-Llama-70B. For example, Llama-3.3-70B-Instruct is indistinguishable from Meta-Llama-3-8B ($p \approx 0.996$) and Mistral-7B-Instruct ($p \approx 0.285$), indicating shared behavior across these variants. More details in the Technical Appendix.

Llama-3.2-3B vs Llama-3.2-3B-Instruct are statistically different ($p \approx 0.0436$), proving the effect of instruction tuning. Moreover, Llama-3.1-8B models (with or without instruction tuning) differ meaningfully from Llama-3.3-70B-Instruct ($p$ ranging from $2 \times 10^{-4}$ to $3 \times 10^{-5}$), showing a shift in behavior as model size increases, even when other models at 7B–8B scale remain similar. We also note a statistically significant difference between Mistral-7B-Instruct and Mixtral-8x7B-Instruct ($p \approx 0.015$), suggesting that the sparse MoE architecture introduces measurable behavioral changes relative to its dense counterpart.

For reasons as to model failures, we observe patterns tied to architectural and training nuances. For example, deepSeek models, despite leading in detection tasks (e.g., DS-R1D-L70B at 49.57\% $C_{detect}$), falter in extraction. In contrast, Mixture-of-Experts (MoE) architectures like Mixtral-8x7B-I-0.1 offer advantages in selective activation, enabling better handling of complex extractions (e.g., 68.06\% on $X_{0,1,1}$), as they dynamically route queries to specialized ``experts" for nuanced causal parsing. These insights suggest that failures stem not just from data but from how models manage context and computation, pointing to opportunities like context-window extensions or MoE-inspired fine-tuning for improved PCD.

    \begin{table*}[ht]
    \centering
    \scriptsize
    \setlength{\tabcolsep}{3pt}
    Table 7: {Bucketization of Errors for Qwen2.5-C-32B-I on Extraction Tasks.}
    \begin{tabular*}{\linewidth}{@{\extracolsep{\fill}} lccccccccccc }
    \toprule
    Incorrect & Halluc. data & Switch CE & False Neg & False Pos & New sent. & Rewrite CE & Mixed & Others & Missing & Correct \\
    \midrule
    6.07\% & 4.47\% & 4.06\% & 3.22\% & 0.31\% & 0.10\% & 0.10\% & 0.14\% & 0.28\% & 35.70\% & 45.55\% \\
    \bottomrule
    \end{tabular*}
    
    \label{tab:error-distribution}
    \end{table*}

\subsubsection{Inter-sentential difficulty}
When the cause--effect pair spans sentences ($B=1$), performance without explicit markers drops sharply for most models, this is especially concerning for biomedical applications. Models' struggles on tasks 2c and 2g, which Figure 4 shows are heavily composed of "General Domain \& News" and "Multi-Domain / Abstract" text (categories which also contain healthcare-related text), highlight a significant bottleneck in their ability to synthesize information from dense research articles. This limitation could hinder automated evidence extraction for systematic reviews or drug discovery pipelines, where critical findings often connect ideas across multiple sentences. With the presence of causal markers like `causes' ($M=1$), the gap narrows or reverses (DS-R1D-L70B 50.64\% vs.\ 47.66\%), indicating heavy reliance on explicit cues rather than robust discourse reasoning. This is also the reason why they fail Task 1a, because the False class is not true causality, but it has causal markers. Inter-sentential cases show greater variance and generally lower values, especially with multiple pairs ($r>1$). Without markers, $X_{0,1,>1}$ is often single-digits (e.g., 3--8\% for many models). Markers help but do not close the gap for most models; only a few strong instruction-tuned models reach mid-30s for $X_{1,1,>1}$. Even for single-pair extraction, many models fall from intra ($X_{,0,1}$) to inter ($X_{,1,1}$) unless highly instruction-tuned (contrast DS-R1D-Q32B 29.73\% $\to$ 30.82\% vs.\ L3.1-8B-I 42.67\% $\to$ 28.97\%). 

This struggle with implicit and inter-sentential relations directly impacts potential biomedical applications. For instance, in EHR analysis, a causal link like ``Patient's hypertension led to renal complications" might be described across several sentences without explicit markers. The low performance on tasks representing this scenario indicates that current models would likely miss such critical insights, potentially delaying diagnoses or interventions. By benchmarking across domains, we show that while some models excel in explicit biomedical extractions, they must be specifically improved to handle the implicit nature of clinical text to become truly effective healthcare tools.

\begin{enumerate}
  \item Inter-sentential PCD is notably harder without markers.
  \item LLMs rely heavily on explicit markers, this makes detection harder when such markers are absent, but when present, they help the models extract causal pairs more reliably.
  \item Multi-pair, inter-sentential extraction remains a clear failure point for these models, suggesting a need for better cross-sentence reasoning, and connective modeling.
\end{enumerate}

LLMs struggling with implicit and inter-sentential causal relations have direct implications for biomedical tasks. For instance, in EHR analysis, where causal links like ``patient's hypertension led to renal complications" might span multiple sentences without explicit markers, such low performance could miss critical insights, potentially delaying diagnoses or interventions. By benchmarking across domains, we highlight how models like Qwen2.5-C-32B-I, which excel in explicit biomedical extractions, could be fine-tuned to bridge this gap, ultimately enhancing tools for clinical decision-making.

\subsubsection{Error analysis}
Table 7 is dominated by missing relations (35.7\%) and having 45.55\% (unweighted average) accurate extractions is promising, knowing this method is totally unsupervised. The main actionable error sources are Incorrect (6.07\%), Hallucinated data (4.47\%), and Switch CE (4.06\%), while false positives and the other categories are rare. Improving recall, through avenues like a richer causal phrasing augmentation and targeted fine-tuning, alongside mechanisms to catch and correct swapped cause/effect pairs and to suppress hallucinated content would yield the biggest results. To improve performance, future work would focus on boosting recall (reducing the ``missing” values) without substantially degrading the low false-positive rate—e.g., through (1) better pattern coverage, (2) data augmentation for underrepresented causal connections, and (3) refining the model’s sensitivity to causal direction (to reduce Switch CE errors).  Moreover, there should be targeted fine-tuning on examples that currently fall into the “Incorrect” and “Hallucinated data” buckets.

\section{Limitations}
\begin{itemize}
  \item \textbf{Data curation and sampling:} Heuristic identification of inter- vs.\ intra-sentential spans and restricted validation sampling can introduce selection bias and may not reflect the full variety of causal expressions.
  \item \textbf{Annotation ambiguity:} Although eight human annotators achieved high agreement ($\kappa \ge 0.758$), subjective interpretation and inherent bias in causal labeling remain potential error sources.
  \item \textbf{Scope:} The benchmark targets English-language PCD on open-source models; proprietary systems and cross-lingual scenarios may behave differently.

  \item \textbf{Prompt and threshold sensitivity:} Though we ran multiple prompts, and chose the best results, the overall results still depend on specific prompt formulations and scoring thresholds, which may hinder generalization to other domains.
\end{itemize}

% Beyond the noted constraints, our study is limited by dataset biases; all corpora are English-only, potentially overlooking multilingual causal expressions common in global biomedical literature, and web-mined sources (e.g., CauseNet) introduce noise like outdated or speculative claims that could skew model training. Additionally, our prompting strategies show sensitivity to hyperparameters, such as temperature=0.7, which balances creativity and determinism but may amplify hallucinations in edge cases. This highlights how slight variations could alter outcomes. These factors underscore the need for broader, cleaner datasets and adaptive prompting in future work to enhance the ability for generalization.

\section{Conclusion}
We introduced a unified benchmark to assess LLM proficiency in pairwise causal discovery, a foundational capability for their application in healthcare. Our findings reveal a critical gap: while models show promise on simple, explicit, intra-sentential statements, their effectiveness plummets on the implicit and multi-sentence reasoning required to understand complex clinical narratives and biomedical literature. This trade-off, where no single model excels at both detection and extraction, highlights a significant bottleneck. For LLMs to be safely deployed in medicine, they must be able to reliably identify causal links from patient records (e.g., adverse drug events) or trial data, not just simple text. These findings motivate a clear path forward: future work must prioritize fine-tuning on domain-specific causal corpora (like EHRs and abstracts) and integrating medical knowledge graphs to build models truly capable of supporting clinical decisions and accelerating research. 
 
\section{Acknowledgments}
We thank  Ifeoluwa Kunle-John for assisting with annotation recruitment and Mehidi Mahmud Kaushik for creating the diagrams.

\bibliography{ieee}
\bibliographystyle{IEEEtran}

\section*{Appendix}

\begin{table*}[t]
    \setlength{\tabcolsep}{1mm}
  \centering
  \caption{Experiment 1a Results N=729)}
  \label{tab:exp1-results}
  {\fontsize{9pt}{10}\selectfont
\begin{tabular}{|p{1.5em}|p{7em}|p{5em}|*{9}{>{\centering\arraybackslash}p{3.8em}|}}
    \hline
    S/N & Model & Prompt Style & Full Acc. & Full Prec. & Full Rec. & Full F1 & Attempted N & Att. Acc. & Att. Prec. & Att. Rec. & Att. F1 \\
    \hline
    1 & DeepSeek-R1-0528-Qwen3-8B & GIP & 0.0604 & 0.6429 & 0.0707 & 0.1274 & 74 & 0.5946 & 0.6429 & 0.6429 & 0.6429 \\
     &  & COT & 0.0658 & 0.5778 & 0.0681 & 0.1218 & 78 & 0.6154 & 0.5778 & 0.7027 & 0.6341 \\
     &  & FICL & $-$ & $-$ & $-$ & $-$ & $-$ & $-$ & $-$ & $-$ & $-$ \\
     &  & COT+FICL & $-$ & $-$ & $-$ & $-$ & $-$ & $-$ & $-$ & $-$ & $-$ \\
    \hline
    2 & DeepSeek-R1-Distill-Llama-70B & GIP & 0.6049 & 0.6052 & 0.9110 & 0.7273 & 689 & 0.6401 & 0.6052 & 0.9431 & 0.7373 \\
     &  & COT & 0.6516 & 0.6389 & 0.9031 & 0.7484 & 692 & 0.6864 & 0.6389 & 0.9401 & 0.7607 \\
     &  & FICL & 0.5610 & 0.5985 & 0.8508 & 0.7027 & 646 & 0.6331 & 0.5985 & 0.9448 & 0.7328 \\
\hline
  \end{tabular}
  }
\end{table*}

\begin{table*}[t]
    \setlength{\tabcolsep}{1mm}
  \centering
  \caption{Experiment 1b Results N=321)}
  \label{tab:exp1-results}
  {\fontsize{9pt}{10}\selectfont
  \begin{tabular}{|p{1.5em}|p{7em}|p{5em}|*{9}{>{\centering\arraybackslash}p{3.8em}|}}
  \hline
    S/N & Model & Prompt Style & Full Acc. & Full Prec. & Full Rec. & Full F1 & Attempted N & Att. Acc. & Att. Prec. & Att. Rec. & Att. F1 \\
    \hline
1 & DeepSeek-R1-0528-Qwen3-8B & GIP & 0.0841 & 0.9091 & 0.1010 & 0.1818 & 32 & 0.8438 & 0.9091 & 0.7143 & 0.7999 \\
  &                             & COT                & 0.0717 & 0.6667 & 0.0404 & 0.0762 & 28 & 0.8214 & 0.6667 & 0.5714 & 0.6154 \\
  &                             & FICL               & $-$    & $-$    & $-$    & $-$    & $-$ & $-$    & $-$    & $-$    & $-$ \\
  &                             & COT+FICL           & $-$    & $-$    & $-$    & $-$    & $-$ & $-$    & $-$    & $-$    & $-$ \\
\hline
2 & DeepSeek-R1-Distill-Llama-70B & GIP & 0.7227 & 0.5494 & 0.8990 & 0.6820 & 309 & 0.7508 & 0.5494 & 0.9570 & 0.6980 \\
  &                                & COT                & 0.7570 & 0.5874 & 0.8485 & 0.6942 & 315 & 0.7714 & 0.5874 & 0.8660 & 0.7000 \\
  &                                & FICL               & 0.7321 & 0.5822 & 0.8586 & 0.6939 & 300 & 0.7833 & 0.5822 & 0.9551 & 0.7234 \\
\hline
  \end{tabular}
  }
\end{table*}

\begin{table*}[t]
    \setlength{\tabcolsep}{1mm}
  \centering
  \caption{Experiment 1c Results (N=125)}
  \label{tab:exp1-results}
  {\fontsize{9pt}{10}\selectfont
  \begin{tabular}{|p{1.5em}|p{7em}|p{5em}|*{9}{>{\centering\arraybackslash}p{3.8em}|}}
  \hline
    S/N & Model & Prompt Style & Full Acc. & Full Prec. & Full Rec. & Full F1 & Attempted N & Att. Acc. & Att. Prec. & Att. Rec. & Att. F1 \\
    \hline
1 & DeepSeek-R1-0528-Qwen3-8B & GIP & 0.0400 & 0.4000 & 0.0222 & 0.0421 & 16 & 0.3125 & 0.4000 & 0.2000 & 0.2667 \\
  &                            & COT                & 0.0320 & 1.0000 & 0.0444 & 0.0851 & 7 & 0.5714 & 1.0000 & 0.5714 & 0.7273 \\
  &                             & FICL               & $-$    & $-$    & $-$    & $-$    & $-$ & $-$    & $-$    & $-$    & $-$ \\
  &                             & COT+FICL           & $-$    & $-$    & $-$    & $-$    & $-$ & $-$    & $-$    & $-$    & $-$ \\
\hline
2 & DeepSeek-R1-Distill-Llama-70B & GIP & 0.7280 & 0.7500 & 0.9333 & 0.8317 & 123 & 0.7398 & 0.7500 & 0.9545 & 0.8400 \\
  &                                & COT                & 0.6720 & 0.7476 & 0.8556 & 0.7979 & 117 & 0.7179 & 0.7476 & 0.9167 & 0.8235 \\
  &                                & FICL               & 0.6640 & 0.7339 & 0.8889 & 0.8040 & 118 & 0.7034 & 0.7339 & 0.9302 & 0.8205 \\
\hline
      \end{tabular}
  }
\end{table*}

\begin{table*}[t]
    \setlength{\tabcolsep}{1mm}
  \centering
  \caption{Experiment 1d Results (N=43)}
  \label{tab:exp1-results}
  {\fontsize{9pt}{10}\selectfont
  \begin{tabular}{|p{1.5em}|p{7em}|p{5em}|*{9}{>{\centering\arraybackslash}p{3.8em}|}}
  \hline
    S/N & Model & Prompt Style & Full Acc. & Full Prec. & Full Rec. & Full F1 & Attempted N & Att. Acc. & Att. Prec. & Att. Rec. & Att. F1 \\
    \hline
    1 & DeepSeek-R1-0528-Qwen3-8B & GIP & 0.0465 & 1.0000 & 0.0476 & 0.0909 & 3 & 0.6667 & 1.0000 & 0.5000 & 0.6667 \\
  &                            & COT                & 0.0698 & 1.0000 & 0.0476 & 0.0909 & 3 & 1.0000 & 1.0000 & 1.0000 & 1.0000 \\
  &                             & FICL               & $-$    & $-$    & $-$    & $-$    & $-$ & $-$    & $-$    & $-$    & $-$ \\
  &                             & COT+FICL           & $-$    & $-$    & $-$    & $-$    & $-$ & $-$    & $-$    & $-$    & $-$ \\
\hline
2 & DeepSeek-R1-Distill-Llama-70B & GIP & 0.8140 & 0.7500 & 1.0000 & 0.8571 & 42 & 0.8333 & 0.7500 & 1.0000 & 0.8571 \\
  &                               & COT                & 0.8372 & 0.8636 & 0.9048 & 0.8837 & 40 & 0.9000 & 0.8636 & 0.9500 & 0.9048 \\
  &                               & FICL               & 0.6047 & 0.6400 & 0.7619 & 0.6957 & 35 & 0.7429 & 0.6400 & 1.0000 & 0.7805 \\
\hline

          \end{tabular}
  }
\end{table*}

\begin{table*}[t]
\centering
\caption{Performance Metrics Across Models and Prompting Methods}
\label{tab:aaai_perf}
\begin{tabular}{@{} p{3.5cm} l r r r r r r r r@{}}
\toprule
Model & Method & S.S & Acc & Recall & F$_1$ Score & Att. S.S Size & Att. Acc & Att. Recall & Att. F$_1$ Score \\
\midrule
DeepSeek-R1-0528-Qwen3-8B & CoT & 2586 & 0.0027 & 0.0027 & 0.0054 & 24 & 0.2917 & 0.2917 & 0.4516 \\
                          & FICL & 2536 & 0.0233 & 0.0233 & 0.0455 & 176 & 0.3352 & 0.3352 & 0.5021 \\
                          & L2M  & 2476 & 0.0735 & 0.0735 & 0.1369 & 445 & 0.4090 & 0.4090 & 0.5805 \\
                          & RCT  & 2529 & 0.0269 & 0.0269 & 0.0524 & 208 & 0.3269 & 0.3269 & 0.4928 \\
                          & GIP  & 2577 & 0.0074 & 0.0074 & 0.0146 & 130 & 0.1462 & 0.1462 & 0.2550 \\ \hline
\addlinespace
DeepSeek-R1-Distill-Llama-70B & CoT & 2536 & 0.0268 & 0.0268 & 0.0522 & 114 & 0.5965 & 0.5965 & 0.7473 \\
                          & FICL & 2581 & 0.0019 & 0.0019 & 0.0039 & 6 & 0.8333 & 0.8333 & 0.9091 \\
\bottomrule
\end{tabular}
\end{table*}

\onecolumn
\setlength{\LTleft}{\fill}
\setlength{\LTright}{\fill}
\begin{longtable}{p{3.5cm} l r r r r r r r r}
  % first‐page caption and header
  \caption{Performance Metrics Across Models and Prompting Methods for Experiment 2A}
  \label{tab:aaai_perf} \\ 
  \toprule
  Model & Method & Acc & Recall & F$_1$ Score 
        & Att.\ S.S & Att.\ Acc & Att.\ Recall & Att.\ F$_1$ Score \\
  \midrule
  \endfirsthead

  % second+ caption and header (no \multicolumn)
  \caption[]{Performance Metrics Across Models and Prompting Methods for Experiment 2A (continued)} \\ 
  \toprule
  Model & Method & Acc & Recall & F$_1$ Score 
        & Att.\ S.S & Att.\ Acc & Att.\ Recall & Att.\ F$_1$ Score \\
  \midrule
  \endhead

  % footer if you want a “ ” notice
  \midrule
  \multicolumn{10}{r}{\emph{ }}\\ % <-- you *can* drop this too
  \endfoot

  % final bottom rule
  \bottomrule
DeepSeek-R1-0528-Qwen3-8B & CoT  & 0.0035 & 0.0035 & 0.0069 & 24 & 0.3750 & 0.3750 & 0.5455 \\
                          & FICL & 0.0511 & 0.0511 & 0.0973 & 176 & 0.4886 & 0.4886 & 0.6565 \\
                          & L2M  & 0.1386 & 0.1386 & 0.2435 & 445 & 0.5640 & 0.5640 & 0.7213 \\
                          & RCT  & 0.0596 & 0.0596 & 0.1125 & 208 & 0.4808 & 0.4808 & 0.6494 \\
                          & GIP  & 0.0120 & 0.0120 & 0.0237 & 130 & 0.1769 & 0.1769 & 0.3007 \\ \addlinespace
DeepSeek-R1-Distill-Llama-70B & CoT  & 0.0503 & 0.0503 & 0.0958 & 114 & 0.7105 & 0.7105 & 0.8308 \\
                          & FICL & 0.0043 & 0.0043 & 0.0085 & 6 & 1.0000 & 1.0000 & 1.0000 \\
                          & L2M  & 0.0190 & 0.0190 & 0.0372 & 42 & 0.7381 & 0.7381 & 0.8493 \\
\end{longtable}

\setlength{\LTleft}{\fill}
\setlength{\LTright}{\fill}
\begin{longtable}{p{3.5cm} l r r r r r r r r}
  % first‐page caption and header
  \caption{Performance Metrics Across Models and Prompting Methods for Experiment 2B}
  \label{tab:aaai_perf} \\ 
  \toprule
  Model & Method & Acc & Recall & F$_1$ Score 
        & Att.\ S.S & Att.\ Acc & Att.\ Recall & Att.\ F$_1$ Score \\
  \midrule
  \endfirsthead

  % second+ caption and header (no \multicolumn)
  \caption[]{Performance Metrics Across Models and Prompting Methods for Experiment 2B (continued)} \\ 
  \toprule
  Model & Method & Acc & Recall & F$_1$ Score 
        & Att.\ S.S & Att.\ Acc & Att.\ Recall & Att.\ F$_1$ Score \\
  \midrule
  \endhead

  % footer if you want a “continued on next page” notice
  \midrule
  \endfoot

  % final bottom rule
  \bottomrule
DeepSeek-R1-0528-Qwen3-8B & CoT  & 0.0019 & 0.0019 & 0.0039 & 14 & 0.2143 & 0.2143 & 0.3529 \\
                          & FICL & 0.0351 & 0.0351 & 0.0678 & 125 & 0.4240 & 0.4240 & 0.5955 \\
                          & L2M  & 0.1261 & 0.1261 & 0.2239 & 300 & 0.6067 & 0.6067 & 0.7552 \\
                          & RCT  & 0.0526 & 0.0526 & 0.1000 & 167 & 0.4611 & 0.4611 & 0.6311 \\
                          & GIP  & 0.0071 & 0.0071 & 0.0142 & 65 & 0.1692 & 0.1692 & 0.2895 \\ \addlinespace
DeepSeek-R1-Distill-Llama-70B & CoT  & 0.0539 & 0.0539 & 0.1023 & 95 & 0.7895 & 0.7895 & 0.8824 \\
                          & FICL & 0.0026 & 0.0026 & 0.0052 & 4 & 0.7500 & 0.7500 & 0.8571 \\
                          & L2M  & 0.0117 & 0.0117 & 0.0231 & 25 & 0.6000 & 0.6000 & 0.7500 \\
\end{longtable}

\setlength{\LTleft}{\fill}
\setlength{\LTright}{\fill}
\begin{longtable}{p{3.5cm} l r r r r r r r r}
  % first‐page caption and header
  \caption{Performance Metrics Across Models and Prompting Methods for Experiment 2C}
  \label{tab:aaai_perf} \\ 
  \toprule
  Model & Method & Acc & Recall & F$_1$ Score 
        & Att.\ S.S & Att.\ Acc & Att.\ Recall & Att.\ F$_1$ Score \\
  \midrule
  \endfirsthead

  % second+ caption and header (no \multicolumn)
  \caption[]{Performance Metrics Across Models and Prompting Methods for Experiment 2C (continued)} \\ 
  \toprule
  Model & Method & Acc & Recall & F$_1$ Score 
        & Att.\ S.S & Att.\ Acc & Att.\ Recall & Att.\ F$_1$ Score \\
  \midrule
  \endhead

  % footer if you want a “continued on next page” notice
  \midrule
  \endfoot

  % final bottom rule
  \bottomrule

DeepSeek-R1-0528-Qwen3-8B & CoT  & 0.0000 & 0.0000 & 0.0000 & 1 & 0.0000 & 0.0000 & 0.0000 \\
                          & FICL & 0.0471 & 0.0471 & 0.0899 & 5 & 0.8000 & 0.8000 & 0.8889 \\
                          & L2M  & 0.0976 & 0.0976 & 0.1778 & 10 & 0.8000 & 0.8000 & 0.8889 \\
                          & RCT  & 0.0714 & 0.0714 & 0.1333 & 11 & 0.5455 & 0.5455 & 0.7059 \\
                          & GIP  & 0.0000 & 0.0000 & 0.0000 & 0 & 0.0000 & 0.0000 & 0.0000 \\ \addlinespace
DeepSeek-R1-Distill-Llama-70B & CoT  & 0.0118 & 0.0118 & 0.0233 & 2 & 0.5000 & 0.5000 & 0.6667 \\
                          & FICL & 0.0000 & 0.0000 & 0.0000 & 1 & 0.0000 & 0.0000 & 0.0000 \\
                          & L2M  & 0.0000 & 0.0000 & 0.0000 & 0 & 0.0000 & 0.0000 & 0.0000 \\
\end{longtable}

\setlength{\LTleft}{\fill}
\setlength{\LTright}{\fill}
\begin{longtable}{p{3.5cm} l r r r r r r r r}
  % first‐page caption and header
  \caption{Performance Metrics Across Models and Prompting Methods for Experiment 2D}
  \label{tab:aaai_perf} \\ 
  \toprule
  Model & Method & Acc & Recall & F$_1$ Score 
        & Att.\ S.S & Att.\ Acc & Att.\ Recall & Att.\ F$_1$ Score \\
  \midrule
  \endfirsthead

  % second+ caption and header (no \multicolumn)
  \caption[]{Performance Metrics Across Models and Prompting Methods for Experiment 2D (continued)} \\ 
  \toprule
  Model & Method & Acc & Recall & F$_1$ Score 
        & Att.\ S.S & Att.\ Acc & Att.\ Recall & Att.\ F$_1$ Score \\
  \midrule
  \endhead

  % footer if you want a “continued on next page” notice
  \midrule
  \endfoot
DeepSeek-R1-0528-Qwen3-8B & CoT  & 0.0218 & 0.0218 & 0.0426 & 26 & 0.7692 & 0.7692 & 0.8696 \\
                          & FICL & 0.1515 & 0.1515 & 0.2632 & 179 & 0.5978 & 0.5978 & 0.7483 \\
                          & L2M  & 0.3805 & 0.3805 & 0.5513 & 468 & 0.7179 & 0.7179 & 0.8358 \\
                          & RCT  & 0.1973 & 0.1973 & 0.3296 & 255 & 0.7216 & 0.7216 & 0.8383 \\
                          & GIP  & 0.0118 & 0.0118 & 0.0233 & 78 & 0.1538 & 0.1538 & 0.2667 \\ \addlinespace
DeepSeek-R1-Distill-Llama-70B & CoT  & 0.0799 & 0.0799 & 0.1479 & 71 & 0.5493 & 0.5493 & 0.7091 \\
                          & FICL & 0.0036 & 0.0036 & 0.0072 & 4 & 0.2500 & 0.2500 & 0.4000 \\
                          & L2M  & 0.0672 & 0.0672 & 0.1259 & 54 & 0.6667 & 0.6667 & 0.8000 \\
\bottomrule

\end{longtable}

\setlength{\LTleft}{\fill}
\setlength{\LTright}{\fill}
\begin{longtable}{p{3.5cm} l r r r r r r r r}
  % first‐page caption and header
  \caption{Performance Metrics Across Models and Prompting Methods for Experiment 2E}
  \label{tab:aaai_perf} \\ 
  \toprule
  Model & Method & Acc & Recall & F$_1$ Score 
        & Att.\ S.S & Att.\ Acc & Att.\ Recall & Att.\ F$_1$ Score \\
  \midrule
  \endfirsthead

  % second+ caption and header (no \multicolumn)
  \caption[]{Performance Metrics Across Models and Prompting Methods for Experiment 2E (continued)} \\ 
  \toprule
  Model & Method & Acc & Recall & F$_1$ Score 
        & Att.\ S.S & Att.\ Acc & Att.\ Recall & Att.\ F$_1$ Score \\
  \midrule
  \endhead

  % footer if you want a “continued on next page” notice
  \midrule
  \endfoot
DeepSeek-R1-0528-Qwen3-8B & CoT  & 0.0000 & 0.0000 & 0.0000 & 23 & 0.0000 & 0.0000 & 0.0000 \\
                          & FICL & 0.0109 & 0.0109 & 0.0216 & 173 & 0.1040 & 0.1040 & 0.1885 \\
                          & L2M  & 0.0327 & 0.0327 & 0.0634 & 467 & 0.1542 & 0.1542 & 0.2672 \\
                          & RCT  & 0.0156 & 0.0156 & 0.0307 & 229 & 0.1354 & 0.1354 & 0.2385 \\
                          & GIP  & 0.0004 & 0.0004 & 0.0008 & 133 & 0.0075 & 0.0075 & 0.0149 \\ \addlinespace
DeepSeek-R1-Distill-Llama-70B & CoT  & 0.0000 & 0.0000 & 0.0000 & 76 & 0.0000 & 0.0000 & 0.0000 \\
                          & FICL & 0.0008 & 0.0008 & 0.0016 & 5 & 0.4000 & 0.4000 & 0.5714 \\
                          & L2M  & 0.0004 & 0.0004 & 0.0008 & 34 & 0.0294 & 0.0294 & 0.0571 \\
\bottomrule

\end{longtable}

\setlength{\LTleft}{\fill}
\setlength{\LTright}{\fill}
\begin{longtable}{p{3.5cm} l r r r r r r r r}
  % first‐page caption and header
  \caption{Performance Metrics Across Models and Prompting Methods for Experiment 2F}
  \label{tab:aaai_perf} \\ 
  \toprule
  Model & Method & Acc & Recall & F$_1$ Score 
        & Att.\ S.S & Att.\ Acc & Att.\ Recall & Att.\ F$_1$ Score \\
  \midrule
  \endfirsthead

  % second+ caption and header (no \multicolumn)
  \caption[]{Performance Metrics Across Models and Prompting Methods for Experiment 2F (continued)} \\ 
  \toprule
  Model & Method & Acc & Recall & F$_1$ Score 
        & Att.\ S.S & Att.\ Acc & Att.\ Recall & Att.\ F$_1$ Score \\
  \midrule
  \endhead

  % footer if you want a “continued on next page” notice
  \midrule
  \endfoot

DeepSeek-R1-0528-Qwen3-8B & CoT  & 0.0000 & 0.0000 & 0.0000 & 3   & 0.0000 & 0.0000 & 0.0000 \\
                          & FICL & 0.0000 & 0.0000 & 0.0000 & 21  & 0.0000 & 0.0000 & 0.0000 \\
                          & L2M  & 0.0712 & 0.0712 & 0.1330 & 74  & 0.3514 & 0.3514 & 0.5200 \\
                          & RCT  & 0.0275 & 0.0275 & 0.0535 & 45  & 0.2222 & 0.2222 & 0.3636 \\
                          & GIP  & 0.0000 & 0.0000 & 0.0000 & 14  & 0.0000 & 0.0000 & 0.0000 \\
DeepSeek-R1-Distill-Llama-70B & CoT  & 0.0000 & 0.0000 & 0.0000 & 11  & 0.0000 & 0.0000 & 0.0000 \\
                             & FICL & 0.0000 & 0.0000 & 0.0000 & 0   & 0.0000 & 0.0000 & 0.0000 \\
                             & L2M  & 0.0000 & 0.0000 & 0.0000 & 3   & 0.0000 & 0.0000 & 0.0000 \\
\bottomrule

\end{longtable}

\setlength{\LTleft}{\fill}
\setlength{\LTright}{\fill}
\begin{longtable}{p{3.5cm} l r r r r r r r r}
  % first‐page caption and header
  \caption{Performance Metrics Across Models and Prompting Methods for Experiment 2G}
  \label{tab:aaai_perf} \\ 
  \toprule
  Model & Method & Acc & Recall & F$_1$ Score 
        & Att.\ S.S & Att.\ Acc & Att.\ Recall & Att.\ F$_1$ Score \\
  \midrule
  \endfirsthead

  % second+ caption and header (no \multicolumn)
  \caption[]{Performance Metrics Across Models and Prompting Methods for Experiment 2G (continued)} \\ 
  \toprule
  Model & Method & Acc & Recall & F$_1$ Score 
        & Att.\ S.S & Att.\ Acc & Att.\ Recall & Att.\ F$_1$ Score \\
  \midrule
  \endhead

  % footer if you want a “continued on next page” notice
  \midrule
  \endfoot
DeepSeek-R1-0528-Qwen3-8B & CoT  & 0.0000 & 0.0000 & 0.0000 & 2 & 0.0000 & 0.0000 & 0.0000 \\
                          & FICL & 0.0101 & 0.0101 & 0.0199 & 4 & 0.5000 & 0.5000 & 0.6667 \\
                          & L2M  & 0.0150 & 0.0150 & 0.0296 & 16 & 0.1875 & 0.1875 & 0.3158 \\
                          & RCT  & 0.0250 & 0.0250 & 0.0488 & 18 & 0.2778 & 0.2778 & 0.4348 \\
                          & GIP  & 0.0000 & 0.0000 & 0.0000 & 5 & 0.0000 & 0.0000 & 0.0000 \\ \addlinespace
DeepSeek-R1-Distill-Llama-70B & CoT  & 0.0000 & 0.0000 & 0.0000 & 2 & 0.0000 & 0.0000 & 0.0000 \\
                          & FICL & 0.0000 & 0.0000 & 0.0000 & 1 & 0.0000 & 0.0000 & 0.0000 \\
                          & L2M  & 0.0000 & 0.0000 & 0.0000 & 1 & 0.0000 & 0.0000 & 0.0000 \\
\bottomrule

\end{longtable}

\setlength{\LTleft}{\fill}
\setlength{\LTright}{\fill}
\begin{longtable}{p{3.5cm} l r r r r r r r r}
  % first‐page caption and header
  \caption{Performance Metrics Across Models and Prompting Methods for Experiment 2H}
  \label{tab:aaai_perf} \\ 
  \toprule
  Model & Method & Acc & Recall & F$_1$ Score 
        & Att.\ S.S & Att.\ Acc & Att.\ Recall & Att.\ F$_1$ Score \\
  \midrule
  \endfirsthead

  % second+ caption and header (no \multicolumn)
  \caption[]{Performance Metrics Across Models and Prompting Methods for Experiment 2H (continued)} \\ 
  \toprule
  Model & Method & Acc & Recall & F$_1$ Score 
        & Att.\ S.S & Att.\ Acc & Att.\ Recall & Att.\ F$_1$ Score \\
  \midrule
  \endhead

  % footer if you want a “continued on next page” notice
  \midrule
  \endfoot

DeepSeek-R1-0528-Qwen3-8B & CoT  & 0.0000 & 0.0000 & 0.0000 & 2  & 0.0000 & 0.0000 & 0.0000 \\
                          & FICL & 0.0206 & 0.0206 & 0.0403 & 28 & 0.2857 & 0.2857 & 0.4444 \\
                          & L2M  & 0.0566 & 0.0566 & 0.1071 & 39 & 0.5641 & 0.5641 & 0.7213 \\
                          & RCT  & 0.0488 & 0.0488 & 0.0931 & 53 & 0.3585 & 0.3585 & 0.5278 \\
                          & GIP  & 0.0000 & 0.0000 & 0.0000 & 15 & 0.0000 & 0.0000 & 0.0000 \\
DeepSeek-R1-Distill-Llama-70B & CoT  & 0.0000 & 0.0000 & 0.0000 & 9  & 0.0000 & 0.0000 & 0.0000 \\
                          & FICL & 0.0000 & 0.0000 & 0.0000 & 0  & 0.0000 & 0.0000 & 0.0000 \\
                          & L2M  & 0.0000 & 0.0000 & 0.0000 & 1  & 0.0000 & 0.0000 & 0.0000 \\
\bottomrule

\end{longtable}

\begin{table*}[t]
  \centering
  \setlength{\tabcolsep}{4pt}
  \caption{Experiment 2 $C_{\mathrm{extract}}$ (All vs. Att.). Complete Table}
  \label{app:tab_exp2_cextract}
  \begin{tabular}{l *{5}{cc}}
    \toprule
    \textbf{Model}      & \multicolumn{2}{c}{\textbf{CoT}}      & \multicolumn{2}{c}{\textbf{FICL}}     & \multicolumn{2}{c}{\textbf{L2M}}      & \multicolumn{2}{c}{\textbf{RCT}}      & \multicolumn{2}{c}{\textbf{GIP}}     \\
    \cmidrule{2-3} \cmidrule{4-5} \cmidrule{6-7} \cmidrule{8-9} \cmidrule{10-11}
                        & All      & Att.     & All      & Att.     & All      & Att.     & All      & Att.     & All      & Att.    \\
    \midrule
DeepSeek-R1-0528-Qwen3-8B       & 0.34\%   & \textit{16.98\%}    & 4.08\%   & \textit{40.00\%}    & 11.48\%  & \textit{49.32\%}    & 6.22\%   & \textit{40.03\%}    & 0.39\%   & \textit{6.34\%}    \\
DeepSeek-R1-Distill-Llama-70B   & 2.45\%   & \textit{31.87\%}    & 0.14\%   & \textit{30.00\%}    & 1.23\%   & \textit{25.43\%}    & 0.89\%   & \textit{25.01\%}    & 0.48\%   & \textit{12.90\%}   \\
DeepSeek-R1-Distill-Qwen-32B    & 11.38\%  & \textit{34.44\%}    & 24.01\%  & \textit{58.23\%}    & 21.80\%  & \textit{50.32\%}    & 19.02\%  & \textit{41.36\%}    & 17.52\%  & \textit{27.13\%}   \\
Llama-3-8B                      & 3.12\%   & \textit{19.78\%}    & 24.20\%  & \textit{45.62\%}    & 23.38\%  & \textit{38.41\%}    & 11.57\%  & \textit{28.13\%}    & 8.07\%   & \textit{21.65\%}   \\
Llama-3.1-8B-Instruct           & 0.81\%   & \textit{12.96\%}    & 29.44\%  & \textit{60.72\%}    & 29.48\%  & \textit{55.08\%}    & 11.93\%  & \textit{35.41\%}    & 3.86\%   & \textit{25.86\%}   \\
Llama-3.2-3B                    & 0.57\%   & \textit{13.52\%}    & 16.10\%  & \textit{31.22\%}    & 2.22\%   & \textit{20.51\%}    & 2.29\%   & \textit{21.21\%}    & 7.89\%   & \textit{20.22\%}   \\
\bottomrule
  \end{tabular}
\end{table*}

\section{Annotation}
\subsection{Annotation Results}
\begin{table*}[ht]
  \centering
  \small
  \setlength{\tabcolsep}{6pt}

  \begin{subtable}[t]{0.45\linewidth}
    \centering
    \caption{Explicit Intra-sentential Annotation}
    \begin{tabular}{lr}
      \toprule
      Metric & Value \\
      \midrule
      Both raters said 1 & 635 \\
      F = 1, G = 0      & 96  \\
      F = 0, G = 1      & 107 \\
      Both raters said 0 & 563 \\
      \midrule
      Total agreement            & 1401 \\
      Observed agreement         & 0.8551 \\
      Expected agreement (chance)& 0.5013 \\
      Cohen's $\kappa$           & 0.7095 \\
      \bottomrule
    \end{tabular}
  \end{subtable}\hfill
  \begin{subtable}[t]{0.45\linewidth}
    \centering
    \caption{Implicit Intra-sentential Annotation}
    \begin{tabular}{lr}
      \toprule
      Metric & Value \\
      \midrule
      Both raters said 1 & 696 \\
      F = 1, G = 0      & 64  \\
      F = 0, G = 1      & 52  \\
      Both raters said 0 & 201 \\
      \midrule
      Total agreement            & 1013 \\
      Observed agreement         & 0.8855 \\
      Expected agreement (chance)& 0.6193 \\
      Cohen's $\kappa$           & 0.6992 \\
      \bottomrule
    \end{tabular}
  \end{subtable}

  \vspace{1em}

  \begin{subtable}[t]{0.45\linewidth}
    \centering
    \caption{Explicit Inter-sentential Annotation}
    \begin{tabular}{lr}
      \toprule
      Metric & Value \\
      \midrule
      Both raters said 1 & 225 \\
      F = 1, G = 0      & 25  \\
      F = 0, G = 1      & 36  \\
      Both raters said 0 & 164 \\
      \midrule
      Total agreement            & 450    \\
      Observed agreement         & 0.8644 \\
      Expected agreement (chance)& 0.5089 \\
      Cohen's $\kappa$           & 0.7240 \\
      \bottomrule
    \end{tabular}
  \end{subtable}\hfill
  \begin{subtable}[t]{0.45\linewidth}
    \centering
    \caption{Implicit Inter-sentential Annotation}
    \begin{tabular}{lr}
      \toprule
      Metric & Value \\
      \midrule
      Both raters said 1 & 95  \\
      F = 1, G = 0      & 14  \\
      F = 0, G = 1      & 0   \\
      Both raters said 0 & 203 \\
      \midrule
      Total agreement            & 312    \\
      Observed agreement         & 0.9551 \\
      Expected agreement (chance)& 0.5589 \\
      Cohen's $\kappa$           & 0.8983 \\
      \bottomrule
    \end{tabular}
  \end{subtable}

  \caption{Cohen's $\kappa$ and agreement breakdown for the Experiment data.}
  \label{app:tab:cohen_kappa_subtables}
\end{table*}

\subsection{Annotation Policies} \label{app:Annotation_Policies}
\subsubsection*{A. General Annotation Principles}
\begin{enumerate}
  \item \textbf{“Genuine” causality}\\
  A specific event/state “A” brings about a specific event/state “B” in the described world. A causal link is only present if the text states it as a fact.
  \item \textbf{Positive labels only when both spans are clear}\\
  If you cannot unambiguously locate both the cause span and the effect span, treat the item as non‑causal (or skip the question).
  \item \textbf{Span granularity}
  \begin{itemize}
    \item Take the smallest contiguous phrase that fully expresses the cause (respectively effect).
    \item Exclude cue words such as “because,” “therefore,” “leading to,” etc.
  \end{itemize}
  \item \textbf{Negations, modals, or reported speech}\\
  Statements where the effect is negated, such as “Exercise causes not death,” are not considered causal for these tasks. Assumptions/claims (“Scientists claim …”) count as non‑causal. Statements like “He claimed X causes Y” are not causal.
  \item \textbf{Multiple causes or effects}\\
  Never join multiple causes or effects with “and”; output one pair per mapping. For example, if A causes B and C, the pairs are Cause: A, Effect: B and Cause: A, Effect: C.
  \item \textbf{No Abstract References}\\
  Vague statements like “It causes bleeding” are not causal because the cause is not identified in the text.
  \item \textbf{Ambiguity rule}\\
  If you are unsure, label 0 in discovery or leave the JSON array empty in attribution. (If your adjudication workflow needs a “maybe,” use 0.5, but only if agreed team‑wide.)
\end{enumerate}

\subsubsection*{B. Causal Discovery Experiments 1a--1d}

\textbf{Binary judgment: is there a causal link?}

\begin{enumerate}
  \item \textbf{Marked, Single‑Sentence}\\
    \textbf{Objective:} Determine if a single sentence contains a causal relationship explicitly signaled by a marker (e.g., “causes,” “due to,” “because,” “as a result of”).  
    \textbf{Input:} One sentence.  
    \textbf{Task:} Label as 1 (Causal) or 0 (Not Causal).
  \item \textbf{Unmarked, Single‑Sentence}\\
    \textbf{Objective:} Determine if a single sentence implies a causal relationship through world knowledge or logical sequence, without an explicit marker.  
    \textbf{Input:} One sentence.  
    \textbf{Task:} Label as 1 (Causal) or 0 (Not Causal).
  \item \textbf{Marked, Inter‑sentential}\\
    \textbf{Objective:} Determine if a causal link exists across 2–3 sentences, explicitly signaled by a marker (e.g., “As a result,” “Therefore”).  
    \textbf{Input:} 2–3 sentences.  
    \textbf{Task:} Label as 1 (Causal) or 0 (Not Causal).
  \item \textbf{Unmarked, Inter‑sentential}\\
    \textbf{Objective:} Determine if a causal link is implied across 2–3 sentences without an explicit marker.  
    \textbf{Input:} 2–3 sentences.  
    \textbf{Task:} Label as 1 (Causal) or 0 (Not Causal).
\end{enumerate}

\subsubsection*{C. Causal Attribution Experiments 2a--2h}

The goal here is to extract exact text spans for cause and effect.

\begin{enumerate}
  \item \textbf{Explicit, Single‑Sentence, Single Pair}\\
    \textbf{Objective:} Extract one clear cause–effect pair from a sentence with an explicit marker.  
    \textbf{Task:} \{\texttt{"Cause": "..."} , \texttt{"Effect": "..."}\}.
  \item \textbf{Implicit, Single‑Sentence, Single Pair}\\
    \textbf{Objective:} Extract one cause–effect pair from a sentence where the link is implicit.  
    \textbf{Task:} \{\texttt{"Cause": "..."} , \texttt{"Effect": "..."}\}.
  \item \textbf{Explicit, Inter‑sentential, Single Pair}\\
    \textbf{Objective:} Extract one pair from 2–3 sentences with an explicit marker.  
    \textbf{Task:} \{\texttt{"Cause": "..."} , \texttt{"Effect": "..."}\}.
  \item \textbf{Implicit, Inter‑sentential, Single Pair}\\
    \textbf{Objective:} Extract one pair from 2–3 sentences where the link is implicit.  
    \textbf{Task:} \{\texttt{"Cause": "..."} , \texttt{"Effect": "..."}\}.
  \item \textbf{Explicit, Single‑Sentence, Multiple Pairs}\\
    \textbf{Objective:} Extract all distinct pairs from a sentence with explicit markers (one pair per cause–effect).  
    \textbf{Task:} List of \{\texttt{"Cause": "..."} , \texttt{"Effect": "..."}\}.
  \item \textbf{Implicit, Single‑Sentence, Multiple Pairs}\\
    \textbf{Objective:} Extract all distinct implicit pairs from one sentence.  
    \textbf{Task:} List of \{\texttt{"Cause": "..."} , \texttt{"Effect": "..."}\}.
  \item \textbf{Explicit, Inter‑sentential, Multiple Pairs}\\
    \textbf{Objective:} Extract all distinct pairs from 2–3 sentences with explicit markers.  
    \textbf{Task:} List of \{\texttt{"Cause": "..."} , \texttt{"Effect": "..."}\}.
  \item \textbf{Implicit, Inter‑sentential, Multiple Pairs}\\
    \textbf{Objective:} Extract all distinct implicit pairs from 2–3 sentences.  
    \textbf{Task:} List of \{\texttt{"Cause": "..."} , \texttt{"Effect": "..."}\}.
\end{enumerate}

\addcontentsline{toc}{section}{Appendix: Call for Annotators}


# Call for Annotators

We're seeking meticulous annotators to support our causal discovery experiments. You'll read sentences (and sentence pairs) and mark whether and how a cause–effect relationship is expressed. Strong attention to detail, consistency in labeling, and familiarity with explicit vs. implicit causality are key.

We have placed safeguards to make sure those that use AI to answer this form will be filtered out. The major test questions have been designed to stump premium AI models like ChatGPT 4.5 and Gemini 2.5 Pro. Therefore, the use of AI would be to your own disadvantage.

Only successful candidates would be contacted.

This form has 3 sections of assessments (5 questions, 5 questions then 10 questions).

* Indicates required question

1. Email *

2. Name *

3. Email *

4. Link to your CV (ensure you grant access)

## Brief Description

5. **A brief explanation of some key terms**

   **Causal Discovery**
   Identifying whether a causal relationship exists between two items (e.g. concepts, events) in text.

   **Explicit Causality**: signaled by clear cue-words (e.g. "because," "causes," "therefore"). Example: He won the election because he campaigned well.

   **Implicit Causality**: no overt markers—causality is inferred from context or temporal/semantic ordering. Example: He ate too much food so he was congested.

   **Single-Sentence**: both cause and effect occur within the same sentence.

   **Intersentential**: the cause appears in one sentence and the effect in another. Example: She skipped breakfast. As a result, she felt dizzy.

   **Intrasentential causality** refers to a causal relation where both the cause and effect appear within the same sentence—typically in separate clauses—rather than across two sentences. Example: Because the coolant was low, the engine overheated.

   **Some things to note**

   If you can't clearly identify the cause and effect in the passage given then it is probably not causal

   Abstract references do not count. 'It causes bleeding' is not causal because we don't know what causes the bleeding.

   We don't want negative c-e pairs. 'Exercise causes not death' is not causal for our experiment.

   Finally, a claim of causality is not causality. 'He claimed loud volumes cause deafness' is not causal.

   Do you understand?

   *Mark only one oval.*

   ◯ Yes
   ◯ No (Read again)

## Assessment 1

Conceptual Knowledge

6. 1. Which of the following is an example of implicit causality? *

   *Mark only one oval.*

   ◯ Smoking causes lung cancer.
   ◯ The lights went out after the storm.
   ◯ Dogs are mammals

7. 2. Intersentential causality occurs when cause and effect appear in separate sentences. *

   *Mark only one oval.*

   ◯ True
   ◯ False

8. **3. What is "explicit causality"?** *

   *Mark only one oval.*

   ○ A cause–effect relation signaled by a clear cue-word (e.g. "because," "therefore")
   ○ A relation where cause and effect appear in separate sentences
   ○ Any inferred causal link without markers
   ○ A relation with multiple causes

9. **4. Which of these is *not* the clearest causal marker?** *

   *Mark only one oval.*

   ○ because
   ○ therefore
   ○ results in
   ○ leading to

10. **5. What makes a pair "marked" vs. "unmarked"?** *

    *Mark only one oval.*

    ○ Presence vs. absence of an explicit cue-word.
    ○ Whether it's single-sentence vs. intersentential.
    ○ Number of words in the cause span.
    ○ Whether it's explicit or implicit.

## Answers to Assessment 1

1. Which of the following is an example of implicit causality?
   A) "Smoking causes lung cancer."
   **B) "The lights went out after the storm." - Answer**
   C) "Dogs are mammals."

2. Intersentential causality occurs when cause and effect appear in separate sentences.
   **True - Answer**
   False

3. What is "explicit causality"?
   **A. A cause–effect relation signaled by a clear cue-word (e.g. "because," "therefore") - Answer**
   B. A relation where cause and effect appear in separate sentences
   C. Any inferred causal link without markers
   D. A relation with multiple causes

4. 4. Which of these is *not* the clearest causal marker?
   A. because
   **B. therefore - Answer**
   C. results in
   D. consequently

5. What makes a pair "marked" vs. "unmarked"?
   **A. Presence vs. absence of an explicit cue-word. - Answer**
   B. Whether it's single-sentence vs. intersentential.
   C. Number of words in the cause span.
   D. Whether it's explicit or implicit.

## Binary Causality Identification

For each sentence, choose the option that best describes its causal status **and**, if causal, the correct cause/effect.

11. **1. The basement flooded after the heavy rains.** *

    *Mark only one oval.*

    ○ Causal – Cause: heavy rains; Effect: basement flooded
    ○ Causal – Cause: basement flooded; Effect: heavy rains
    ○ Non-causal
    ○ Causal – Cause: basement; Effect: heavy rains

12. **2. She left early, so she caught the first train.** *

    *Mark only one oval.*

    ○ Causal – Cause: she left early; Effect: caught first train
    ○ Non-causal
    ○ Causal – Cause: first train; Effect: she left early
    ○ Causal – Cause: train schedule; Effect: she left

13. **3. The committee approved the proposal.** *

    *Mark only one oval.*

    ○ Non-causal
    ○ Causal – Cause: the committee; Effect: approved the proposal
    ○ Causal – Cause: proposal; Effect: committee approval
    ○ Causal – Cause: approval; Effect: proposal

14. **4. He overslept because his alarm didn't ring.** *

    *Mark only one oval.*

    ○ Causal – Cause: alarm didn't ring; Effect: overslept
    ○ Non-causal
    ○ Causal – Cause: overslept; Effect: alarm didn't ring
    ○ Causal – Cause: morning routine; Effect: oversleeping

15. **5. The traffic jam delayed our lunch plans.** *

    *Mark only one oval.*

    ○ Causal – Cause: traffic jam; Effect: lunch plans delayed
    ○ Non-causal
    ○ Causal – Cause: lunch plans; Effect: traffic jam
    ○ Causal – Cause: delayed plans; Effect: traffic jam

## Answers to Assessment 2

1. The basement flooded after the heavy rains.
   **A. Causal – Cause: heavy rains; Effect: basement flooded - Answer**
   B. Causal – Cause: basement flooded; Effect: heavy rains
   C. Non-causal
   D. Causal – Cause: basement; Effect: heavy rains

2. She left early, so she caught the first train."
   **A. Causal – Cause: she left early; Effect: caught first train - Answer**
   B. Non-causal
   C. Causal – Cause: first train; Effect: she left early
   D. Causal – Cause: train schedule; Effect: she left

3. The committee approved the proposal."
   **A. Non-causal - Answer**
   B. Causal – Cause: the committee; Effect: approved the proposal
   C. Causal – Cause: proposal; Effect: committee approval
   D. Causal – Cause: approval; Effect: proposal

4. He overslept because his alarm didn't ring.
   **A. Causal – Cause: alarm didn't ring; Effect: overslept - Answer**
   B. Non-causal
   C. Causal – Cause: overslept; Effect: alarm didn't ring
   D. Causal – Cause: morning routine; Effect: oversleeping

5. The traffic jam delayed our lunch plans.
   **A. Causal – Cause: traffic jam; Effect: lunch plans delayed - Answer**
   B. Non-causal
   C. Causal – Cause: lunch plans; Effect: traffic jam
   D. Causal – Cause: delayed plans; Effect: traffic jam

## Last Round

16. **1. Since hwax causes not jyka and gyzp, replacing jyka (which positively affects lirg) with its negative form (not jyka) reduces lirg's likelihood, reflecting in the question where the change reduces lirg's occurrence.** *

    *Mark only one oval.*

    ○ Causal
    ○ Non-causal

17. **2. "The traditionalists also claim that progressives are not only building music that steps further and further away from Hip Hop , but causes ignorance and misconceptions between fans of more progressive Hip Hop and fans of traditional Hip Hop culture ."** *

    *Mark only one oval.*

    ○ Causal
    ○ Non-causal

18. **3. They can exist on a small scale, such as an oil field spill catching fire, or on a huge scale, as in geyser-like jets of flames from ignited high pressure wells. A frequent cause of a well fire is a high-pressure blowout during drilling operations.** *

    *Mark only one oval.*

    ○ Intersentential causality
    ○ Intersentential non-causality
    ○ Intrasentential causality
    ○ Intrasentential non-causality

19. **4. What is the cause and the effect contained in this sentence? If there are multiple, list them out:**

    Malfunction in heparin sulfate synthesis causes chondrocytes to rapidly differentiate .

    _____
    _____
    _____
    _____

20. 5. What is the cause and the effect contained in this sentence? If there are multiple, list them out:

Upon binding , formation of the GC / GCR complex causes dissociation of chaperone proteins from the glucocorticoid receptor enabling the GC / GCR complex to translocate inside the nucleus .

_____
_____
_____
_____

21. 6. What is the cause and the effect contained in this sentence? If there are multiple, list them out:

Given that not being yupt causes not jyka and ensures qwiu, and considering that either jyka or qwiu can independently cause kwox, the likelihood of kwox remains even in the absence of yupt primarily due to the presence of qwiu.

_____
_____
_____
_____

22. 7. What is the cause and the effect contained in this sentence? If there are multiple, list them out:

Inventing the AIDS Virus is a 1996 book by molecular biologist Peter Duesberg, in which the author argues that HIV does not cause AIDS. Duesberg contends that HIV is a harmless passenger virus and that AIDS is caused by unrelated factors such as drug abuse, antiretroviral medication, chronic malnutrition, poor sanitation, and hemophilia. The unambiguous scientific consensus is that HIV causes AIDS and that Duesberg's claims are incorrect.

_____
_____
_____
_____

23. 8. He skipped breakfast and felt dizzy by noon. *

*Mark only one oval.*

○ Cause: "He skipped breakfast" | Effect: "felt dizzy by noon"
○ Cause: "felt dizzy" | Effect: "skipped breakfast"
○ Cause: "breakfast and felt" | Effect: "dizzy by noon"
○ Non-causal

24. 9. The medication reduced her pain significantly. *

*Mark only one oval.*

○ Cause: "The medication" | Effect: "reduced her pain significantly"
○ Cause: "her pain" | Effect: "The medication reduced"
○ Cause: "medication reduced" | Effect: "her pain"
○ Non-causal

25. 10. Rising sea levels threaten coastal cities. *

*Mark only one oval.*

○ Cause: "Rising sea levels" | Effect: "threaten coastal cities"
○ Cause: "coastal cities" | Effect: "Rising sea levels"
○ Cause: "levels threaten" | Effect: "coastal cities"
○ Non-causal


\label{app:recruitment_form}

% Requires \usepackage{booktabs}
\begin{table}[ht]
\centering
\begin{tabular}{@{}lll@{}}
\toprule
Model (abbr.) & Best Prompt, $D$ & Best Prompt, $X$ \\
\midrule
DS R1-0528-Q3-8B        & COT              & L2M \\
DS R1D-L70B             & General Instruction & CoT \\
DS R1D-Q32B             & General Instruction & L2M \\
L3-8B                   & FICL             & FICL \\
L3.1-8B-I               & COT              & L2M \\
L3.2-3B                 & FICL             & RCT \\
L3.2-3B-I               & General Instruction & FICL \\
MetaL-3.1-8B            & FICL             & L2M \\
MetaL-3.3-70B-I         & COT+FICL         & RCT \\
Mistral-7B-I v0.3       & FICL             & CoT \\
Mixtral-8x7B-I v0.1     & COT              & CoT \\
Qwen2.5-7B-I            & FICL             & L2M \\
Qwen2.5-Coder-32B-I     & COT+FICL         & L2M \\
\bottomrule
\end{tabular}
\end{table}

\begin{table*}[ht]
  \centering
  \footnotesize
  \setlength{\tabcolsep}{6pt}
  \caption{Annotated Causal Sentences Dataset}
  \label{app:tab:example1}
  \begin{tabular}{
      >{\raggedright\arraybackslash}p{5cm}  % Sentence
      >{\raggedright\arraybackslash}p{2.5cm}  % Source
      c  % Label
      c  % Exp Tag
      >{\raggedright\arraybackslash}p{5cm}  % Reasoning
    }
    \toprule
    \textbf{Sentence} & \textbf{Source} & \textbf{Label} & \textbf{Exp Tag} & \textbf{Reasoning} \\
    \midrule
    In the \textit{Inconvenient Truth}, a documentary done in 2007, Al Gore was claiming that Carbon Dioxide is the main cause of global warming.
      & CNet & 0 & 1a ($M=1$, $B=0$)
      & A causal claim or belief does not show any direct link to causality. \\
    \addlinespace
    Most women believed that breast and cervical cancer caused more deaths, even though lung cancer causes more fatalities than the other two combined, claiming 15,000 lives each year.
      & CNet & 1 & 1a ($M=1$, $B=0$)
      & Even though a belief does not cause causality, this sentence directly shows a fact of cause and effect. \\
    \addlinespace
    ``Each tube is inwardly concave, thus forming a central tube through which moisture is sucked.''
      & MedC & 1 & 1b ($M=0$, $B=0$)
      & Causal link established from the shape of the tube to the moisture effect. \\
    \addlinespace
    Amazonian tribes reclaim the rain forest with the help of GPS navigation tools.
      & SE10 & 0 & 1b ($M=0$, $B=0$)
      & This explains the how and not the why, so it is not causal. \\
    \addlinespace
    The cause remains unclear. However, it is presumed to result from disturbances in the biliary system that accelerate the enterohepatic recycling of bile salts.
      & MedC & 0 & 1c ($M=1$, $B=1$)
      & No clear cause and effect, especially seeing the sentence structure. \\
    \addlinespace
    Cancer was the second leading cause of death. As a result, it claimed 287,840 lives (15.2\%).
      & CNet & 1 & 1c ($M=1$, $B=1$)
      & Clear cause and effect with death taking 287,840 lives. \\
    \addlinespace
    The surfer caught the wave. She paddled her board into the ocean.
      & copa & 0 & 1d ($M=0$, $B=1$)
      & This is an effect-cause situation, which renders it invalid. \\
    \addlinespace
    You determined your priorities and income opportunities. You decided the optimal age to claim benefits.
      & FinC & 1 & 1d ($M=0$, $B=1$)
      & This is a sequential action that resulted from a previous action. \\
    \bottomrule
  \end{tabular}
\end{table*}

\begin{table*}[t]
  \centering
  \setlength{\tabcolsep}{3pt}
  \caption{Example sentences with annotations by experiment tag}
  \label{app:tab:example2}
  \begin{tabular}{r |p{5cm}| p{1cm} |p{1.5cm}| p{4cm} |p{1.5cm}| p{1.2cm}| p{1.2cm}|}
    \toprule
    S/N & Sentence & Source & Intra/Inter & Cause--Effect & Imp/Exp & Multi & ExpTag \\
    \midrule
    1121 & Bruce McManus, supports research into the cause, prevention, screening, diagnosis, treatment, support systems, and palliation for heart disease, as well as a wide range of other conditions associated with the lungs, brain, blood and blood vessels. & COL22 & Intra-sentential & (lungs, wide range of other conditions) & explicit & No & \textbf{2a}: $M=1$, $B=0$, $r=1$ \\ \hline
    5386 & With an intervention that ensures everyone listens to jazz, the probability increases to 62\% for lung cancer, demonstrating a clear increase rather than a decrease in the likelihood of developing lung cancer. & CBTxt & Intra-sentential & (intervention ensures everyone listens to jazz, lung cancer probability increases) & implicit & No & \textbf{2b}: $M=0$, $B=0$, $r=1$ \\ \hline
    8877 & Accidents caused by the main character occur in 38\% of the reviewed picture books. Therefore, injuries emerge as the predominant physical symptoms depicted. & PubM & Inter-sentential & ("accidents caused by the main character occur in 38\% of reviewed picture books", "injuries emerge as the predominant physical symptoms depicted") & explicit & No & \textbf{2c}: $M=1$, $B=1$, $r=1$ \\ \hline
    8099 & The friends' dinnertime conversation turned to politics. A debate erupted. & copa & Inter-sentential & ("The friends' dinnertime conversation turned to politics.", "A debate erupted.") & implicit & No & \textbf{2d}: $M=0$, $B=1$, $r=1$ \\ \hline
    2337 & As he shows, citing 1,147 references, there is now good evidence that this diet is a major cause of the epidemic of obesity, type-2 diabetes, heart disease, stroke, cancer, and various mental problems (such as depression and senile dementia) that afflict an increasing number of Americans. & CNet & Intra-sentential & (diet, cancer), (diet, heart\_disease), (diet, stroke) & explicit & Yes & \textbf{2e}: $M=1$, $B=0$, $r>1$ \\ \hline
    7720 & The efforts of Berthelot and Ruelle to put a little order in this mass of literature led only to poor results, and the later researchers, among them in particular Mrs.~Hammer--Jensen, Tannery, Lagercrantz, von~Lippmann, Reitzenstein, Ruska, Bidez, Festugiere and others, could make clear only few points of detail. & MedC & Intra-sentential & (efforts of berthelot and ruelle to organize the literature., poor results in organizing the literature.), (efforts of later researchers (e.g., hammer-jensen, tannery, etc.), clarification of only a few points of detail.) & implicit & Yes & \textbf{2f}: $M=0$, $B=0$, $r>1$ \\ \hline
    8790 & Among emergency department patients with nontraumatic headache, certain warning signs---including changes in headache pattern, motor deficits, and altered consciousness---strongly predicted intracranial pathology. Recognition of these predictors can guide clinical imaging decisions, potentially preventing missed diagnoses. & PubM & Inter-sentential & (warning signs, predicted pathology), (recognition, guide decisions), (guidance, prevent missed diagnoses) & explicit & Yes & \textbf{2g}: $M=1$, $B=1$, $r>1$ \\ \hline
    7224 & Similar to scenario 47, a country's nationwide initiative to reduce food waste was followed by an improvement in overall food security. Concurrently, there was an increase in sustainable farming practices and advancements in food preservation technology. & CN & Inter-sentential & (initiative to reduce food waste, improvement in overall food security.), (initiative to reduce food waste, an increase in sustainable farming practices and advancements in food preservation technology.) & implicit & Yes & \textbf{2h}: $M=0$, $B=1$, $r>1$ \\ 
    \bottomrule
  \end{tabular}
\end{table*}

\begin{table*}[ht]
  \centering
  \small
  \setlength{\tabcolsep}{3pt}
  \caption{Combined Counts and Dataset Distribution by Tag}
  \label{app:tab:dataset_distribution}
  \begin{tabular}{l*{12}{r}r}
    \toprule
    & \multicolumn{12}{c}{\textbf{Source}} & \textbf{} \\
    \cmidrule(lr){2-13} \cmidrule(lr){14-14}
    \textbf{Tag} & CBTxt & CN & CPrb & CNet & COL22 & Copa & CRSS & ECI & FinC & MedC & PubM & SE10 & \textbf{Total} \\
    \midrule
    1a & 44   & 20  & 0   & 1100 & 3    & 0    & -- & 3    & 6    & 20   & --  & 2   & 1198 \\
    1b & 76   & 93  & 2   & 183  & 13   & 0    & -- & 34   & 164  & 269  & --  & 63  & 897  \\
    1c & 0    & 0   & 43  & 238  & 0    & 0    & -- & 0    & 0    & 108  & --  & 0   & 389  \\
    1d & 0    & 0   & 132 & 42   & 0    & 226  & -- & 0    & 68   & 54   & --  & 2   & 524  \\
    \midrule
    \textit{Total} & 120 & 113 & 177 & 1563 & 16 & 226 & -- & 37 & 238 & 451 & -- & 67 & 3008 \\
    \midrule
    \addlinespace
    2a & --   & --  & 1   & 1580 & 303  & --   & -- & 68   & 61   & 314  & --  & 258 & 2641 \\
    2b & 380  & 26  & 70  & --   & 136  & --   & 2  & 495  & 210  & 169  & --  & 54  & 1542 \\
    2c & 5    & --  & --  & 2    & --   & --   & -- & 1    & --   & --   & 79  & --  & 87   \\
    2d & 31   & 12  & 35  & --   & --   & 1000 & -- & --   & 24   & --   & --  & --  & 1102 \\
    2e & 265  & --  & 263 & 1425 & --   & --   & -- & 65   & 337  & 105  & --  & 141 & 2601 \\
    2f & 38   & 8   & 52  & --   & 17   & --   & -- & 204  & 32   & 8    & --  & 5   & 364  \\
    2g & 69   & --  & 1   & --   & --   & --   & -- & 2    & 1    & --   & 126 & --  & 199  \\
    2h & 18   & 288 & 73  & --   & --   & --   & -- & 1    & 9    & --   & --  & --  & 389  \\
    \midrule
    \textit{Total}  & 120  & 113 & 177 & 1563 & 16   & 226  & -- & 37   & 238  & 451  & --  & 67  & 8925 \\
    \bottomrule
  \end{tabular}
\end{table*}

\lstdefinelanguage{json}{
  basicstyle=\ttfamily\small,
  stringstyle=\color{teal},
  showstringspaces=false,
  morestring=[b]",
  sensitive=true,
  morekeywords={true,false,null},
  keywordstyle=\color{blue}\bfseries,
  breaklines=true,
  literate=
    *{0}{{0}}{1}
     {1}{{1}}{1}
     {2}{{2}}{1}
     {3}{{3}}{1}
     {4}{{4}}{1}
     {5}{{5}}{1}
     {6}{{6}}{1}
     {7}{{7}}{1}
     {8}{{8}}{1}
     {9}{{9}}{1}
     {:}{{:}}{1}
     {,}{{,}}{1}
     {\{}{{\{}}{1}
     {\}}{{\}}}{1}
     {[}{{[}}{1}
     {]}{{]}}{1}
}

\begin{tcolorbox}[
  title=\textbf{1. Experiment 2: Zero shot General Instruction Prompt},
  colback=gray!5,
  colframe=black!75,
  fonttitle=\bfseries,
  breakable
]
You are an expert extractor of cause--effect relations in text. Given any sentence(s), output \textbf{only} a flat JSON object listing all cause--effect pairs, observing these rules:
\begin{enumerate}[label=\arabic*.]
  \item \textbf{Pair splitting}
    \begin{itemize}
      \item \textbf{Always} treat each cause$\to$effect link as its own pair. If one cause leads to multiple effects joined by conjunctions, split into separate pairs.
    \end{itemize}
  \item \textbf{Content}
    \begin{itemize}
      \item \texttt{cause}: text span of the cause.
      \item \texttt{effect}: text span of exactly one effect (carry over any necessary context).
      \item \texttt{causal\_markers}: list of marker words/phrases.
      \item \texttt{explicitness}: ``explicit'' or ``implicit''.
      \item \texttt{sentential\_scope}: ``intrasentential'' or ``intersentential''.
    \end{itemize}
\end{enumerate}
Subsequent pairs: append \_2, \_3, etc.\ to each key. Now extract the cause--effect pairs from the following sentence:

\textbf{Input:} \{input\_sentence\}

Use this format:
\begin{verbatim}
{
  "cause": "...",
  "effect": "...",
  "causal_markers": ["..."],
  "explicitness": "...",
  "sentential_scope": "..."
}
\end{verbatim}
\end{tcolorbox}

\begin{tcolorbox}[
  title=\textbf{2.  Experiment 2: Few-Shot Incontext Learning Prompt},
  colback=gray!5,
  colframe=black!75,
  fonttitle=\bfseries,
  breakable
]
You are an expert extractor of cause--effect relations in text. Given any sentence(s), output \textbf{only} a flat JSON object listing all cause--effect pairs, observing these rules:
\begin{enumerate}[label=\arabic*.]
  \item \textbf{Pair splitting}
    \begin{itemize}
      \item \textbf{Always} treat each cause$\to$effect link as its own pair. If one cause leads to multiple effects joined by conjunctions, split into separate pairs.
    \end{itemize}
  \item \textbf{Content}
    \begin{itemize}
      \item \texttt{cause}: text span of the cause.
      \item \texttt{effect}: text span of exactly one effect (carry over any necessary context).
      \item \texttt{causal\_markers}: list of marker words/phrases.
      \item \texttt{explicitness}: ``explicit'' or ``implicit''.
      \item \texttt{sentential\_scope}: ``intrasentential'' or ``intersentential''.
    \end{itemize}
\end{enumerate}
Subsequent pairs: append \_2, \_3, etc.\ to each key.

\textbf{Example 1:}

\textbf{Input:} He heard the loud thunder and went deaf

\textbf{Output:}
\begin{lstlisting}[language=JSON]
{
  "cause": "He heard the loud thunder",
  "effect": "went deaf",
  "causal_markers": [],
  "explicitness": "implicit",
  "sentential_scope": "intrasentential"
}
\end{lstlisting}

\textbf{Example 2:}

\textbf{Input:} The term includes impairments caused by congenital anomaly (e.g., clubfoot, absence of some member, etc.), impairments caused by disease (e.g., poliomyelitis, bone tuberculosis, etc.), and impairments from other causes (e.g., cerebral palsy, amputations, and fractures or burns that cause contractures) * (12) Traumatic brain injury means an acquired injury to the brain caused by an external physical force, resulting in total or partial functional disability or psychosocial impairment, or both, that adversely affects a child's educational performance.

\textbf{Output:}
\begin{lstlisting}[language=JSON]
{
  "cause": "congenital anomaly",
  "effect": "impairments",
  "causal_markers": ["caused by"],
  "explicitness": "explicit",
  "sentential_scope": "intrasentential",

  "cause_2": "disease",
  "effect_2": "impairments",
  "causal_markers_2": ["caused by"],
  "explicitness_2": "explicit",
  "sentential_scope_2": "intrasentential",

  "cause_3": "other causes (e.g., cerebral palsy, amputations, fractures or burns that cause contractures)",
  "effect_3": "impairments",
  "causal_markers_3": ["from"],
  "explicitness_3": "explicit",
  "sentential_scope_3": "intrasentential",

  "cause_4": "fractures or burns",
  "effect_4": "contractures",
  "causal_markers_4": ["cause"],
  "explicitness_4": "explicit",
  "sentential_scope_4": "intrasentential",

  "cause_5": "external physical force",
  "effect_5": "acquired injury to the brain",
  "causal_markers_5": ["caused by"],
  "explicitness_5": "explicit",
  "sentential_scope_5": "intrasentential",

  "cause_6": "acquired injury to the brain",
  "effect_6": "total or partial functional disability",
  "causal_markers_6": ["resulting in"],
  "explicitness_6": "explicit",
  "sentential_scope_6": "intrasentential",

  "cause_7": "acquired injury to the brain",
  "effect_7": "psychosocial impairment",
  "causal_markers_7": ["resulting in"],
  "explicitness_7": "explicit",
  "sentential_scope_7": "intrasentential",

  "cause_8": "total or partial functional disability",
  "effect_8": "adverse effect on a child's educational performance",
  "causal_markers_8": ["adversely affects"],
  "explicitness_8": "explicit",
  "sentential_scope_8": "intrasentential",

  "cause_9": "psychosocial impairment",
  "effect_9": "adverse effect on a child's educational performance",
  "causal_markers_9": ["adversely affects"],
  "explicitness_9": "explicit",
  "sentential_scope_9": "intrasentential"
}
\end{lstlisting}

Now extract the cause--effect pairs from the following sentence:

\textbf{Input:} \{input\_sentence\}

Use this format:
\begin{verbatim}
{
  ... (flat JSON with _2, _3 suffixes as needed) ...
}
\end{verbatim}
\end{tcolorbox}

\begin{tcolorbox}[
  title=\textbf{3.  Experiment 2: Chain-of-thought Prompting},
  colback=gray!5,
  colframe=black!75,
  fonttitle=\bfseries,
  breakable
]
Extract every cause--effect link from the provided input.\\
Think (silently)\\
When a single cause leads to multiple effects or a single effect results from multiple causes---including cases where causes or effects are linked by conjunctions (e.g., ``and,'' ``or,'' ``but'')---split each into separate cause--effect pairs.\\
For each pair, fill in: cause, effect, causal\_markers, explicitness (implicit or explicit), sentential\_scope (intersentential or intrasentential)\\
Answer (only this, no explanations)\\
Return a flat JSON object. Use these keys for the first pair and append \_2, \_3, \dots\ for any additional pairs, all inside the same set of braces:
\begin{verbatim}
{
  "cause": "...",
  "effect": "...",
  "causal_markers": ["..."],
  "explicitness": "...",
  "sentential_scope": "..."
}
\end{verbatim}
Now extract the cause--effect pairs from the following sentence(s):

\textbf{Input:} \{input\_sentence\}
\end{tcolorbox}

\begin{tcolorbox}[
  title=\textbf{4.  Experiment 2: Least to Most Prompt},
  colback=gray!5,
  colframe=black!75,
  fonttitle=\bfseries,
  breakable
]
List the obvious causal markers you spot in the passage (e.g., because, therefore, so, since, as a result).\\
Pair them so that each cause links to exactly one effect.\\
If a single cause is followed by several effects or several causes converge on one effect---including cases where the causes/effects are joined by conjunctions such as ``and,'' ``or,'' ``but''---split them into separate pairs.\\
For each pair, record:
\begin{itemize}
  \item cause
  \item effect
  \item causal\_markers --- list the explicit cue words (use [] if none)
  \item explicitness --- ``explicit'' if a marker is present, otherwise ``implicit''
  \item sentential\_scope --- ``intrasentential'' if cause and effect are in the same sentence, otherwise ``intersentential''
\end{itemize}
Compile a single flat JSON object with these keys for the first pair, then append \_2, \_3, \dots\ for subsequent pairs, all inside the same pair of braces:
\begin{verbatim}
{
  "cause": "...",
  "effect": "...",
  "causal_markers": ["..."],
  "explicitness": "...",
  "sentential_scope": "...",
  "cause_2": "...",
  "effect_2": "...",
  "causal_markers_2": ["..."],
  "explicitness_2": "...",
  "sentential_scope_2": "..."
}
\end{verbatim}
Return only the JSON object produced in step 6---no additional text.\\
Now extract the cause--effect pairs from the following sentence:

\textbf{Input:} \{input\_sentence\}
\end{tcolorbox}

\begin{tcolorbox}[
  title=\textbf{5.  Experiment 2: ReAct Style},
  colback=gray!5,
  colframe=black!75,
  fonttitle=\bfseries,
  breakable
]
You are an expert extractor of cause--effect relations in text.\\
\textbf{Reason:}
\begin{enumerate}[label=\arabic*.]
  \item Read the sentence.
  \item Detect every cause phrase and every directly linked effect phrase.
  \item For each cause$\to$effect link, create a separate pair. 
    \begin{itemize}
      \item If one cause is tied to multiple effects joined by ``and,'' ``or,'' commas, etc., split into distinct pairs.
      \item If multiple causes jointly lead to one effect, split into distinct pairs (duplicate the effect text).
    \end{itemize}
  \item Record any explicit causal markers (e.g., ``because'', ``so'', ``leads to'', ``caused by'', ``resulting in'', ``that adversely affects'').
  \item Decide for each pair whether the causality is explicit or implicit, and whether it is intra- or inter-sentential.
\end{enumerate}
\textbf{Act:} Output \textbf{only} a single flat JSON object \{\dots\}, not an array, with one block containing all pairs. Use these keys for the first pair: 
\begin{verbatim}
"cause": "...", "effect": "...", "causal_markers": ["...", "..."], "explicitness": "explicit" or "implicit", "sentential_scope": "intrasentential" or "intersentential"
\end{verbatim}
For each additional pair, append \_2, \_3, etc., to each key (e.g.,
\begin{verbatim}
"cause_2": "...", "effect_2": "...", "causal_markers_2": ["..."], "explicitness_2": "...", "sentential_scope_2": "..."
\end{verbatim}
).\\
Now extract the cause--effect pairs from the following sentence:

\textbf{Input:} \{input\_sentence\}
\end{tcolorbox}

\end{document}